\documentclass[10pt,twocolumn,letterpaper]{article}

\usepackage[pagenumbers]{iccv} % To force page numbers, e.g. for an arXiv version

\usepackage[dvipsnames]{xcolor}

\definecolor{iccvblue}{rgb}{0.21,0.49,0.74}
\usepackage{wasysym}
\usepackage[pagebackref,breaklinks,colorlinks,allcolors=iccvblue]{hyperref}

\usepackage{tabularx}
\usepackage{graphicx}
\usepackage{amsmath} 
\usepackage{multirow} 
\usepackage{pifont} 
\usepackage{subcaption}
\usepackage{graphicx}
\usepackage{wrapfig}

\usepackage{enumitem}
\usepackage{colortbl}
\usepackage{tablefootnote}
\usepackage{hyperref}

\usepackage{url}
\usepackage[T1]{fontenc}    % use 8-bit T1 fonts
\usepackage{hyperref}       % hyperlinks
\usepackage{url}            % simple URL typesetting
\usepackage{booktabs}       % professional-quality tables
\usepackage{amsfonts}       % blackboard math symbols
\usepackage{nicefrac}       % compact symbols for 1/2, etc.
\usepackage{microtype}      % microtypography
\usepackage{xcolor}         % colors
\usepackage{subcaption} 
\def\paperID{10245} % *** Enter the Paper ID here
\def\confName{ICCV}
\def\confYear{2025}

\newcommand{\ieno}{\textit{i.e.}}
\newcommand{\egno}{\textit{e.g.}}

\title{Q-Adapt: Adapting LMM for Visual Quality Assessment with Progressive Instruction Tuning}
\author{
Yiting Lu$^{1}$, Xin Li$^{1,\textrm{\ding{41}}}$, Haoning Wu$^{2}$, Bingchen Li$^{1}$, Weisi Lin$^{2}$, Zhibo Chen$^{1}$ \\ [1mm]
\textsuperscript{\rm 1} University of Science and Technology of China, 
\textsuperscript{\rm 2} Nanyang Technological University  \\
{\tt \small \{luyt31415,lbc31415926\}@mail.ustc.edu.cn,}
{\tt \small haoning001@e.ntu.edu.sg,} \\
{\tt \small wslin@ntu.edu.sg,} 
{\tt \small \{xin.li, chenzhibo\}@ustc.edu.cn}}

\begin{document}

\twocolumn[{%
\renewcommand\twocolumn[1][]{#1}%
\maketitle

\begin{center}
    \centering
    \vspace{-10pt}
    \captionsetup{type=figure}
    \includegraphics[width=\textwidth]{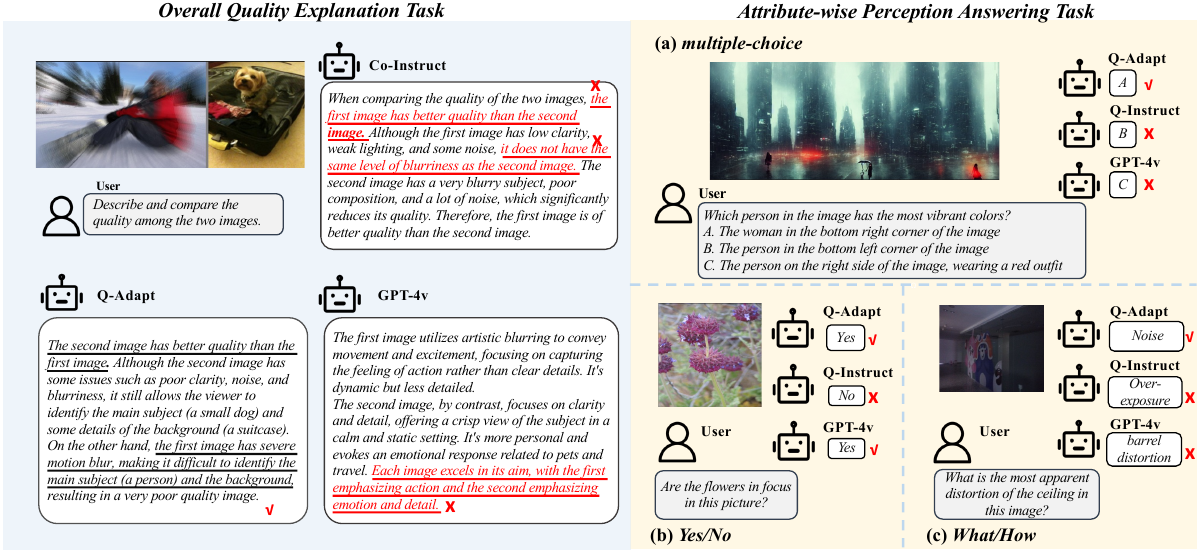}
    \caption{The comparison between existing LMMs and our proposed Q-Adapt on two EIQA tasks (\ieno, the overall quality explanation task, and the attribute-wise perception answering task). Our proposed Q-Adapt can generate more accurate response, benefiting from the reduction of task conflicts and the enhanced synergy between the two tasks, achieved through progressive instruction tuning.}
    \label{fig:task}
\end{center}
}]

\maketitle
\renewcommand{\thefootnote}{}
\footnotetext{
\textsuperscript{\ding{41}} Corresponding author.
}

\begin{abstract}
The rapid advancement of Large Multi-modal Foundation Models (LMM) has paved the way for the possible Explainable Image Quality Assessment (EIQA) with instruction tuning from two perspectives: overall quality explanation, and attribute-wise perception answering. However, existing works usually overlooked the conflicts between these two types of perception explanations during joint instruction tuning, leading to insufficient perception understanding. To mitigate this, we propose a new paradigm for perception-oriented instruction tuning, \ieno, Q-Adapt, which aims to eliminate the conflicts and achieve the synergy between these two EIQA tasks when adapting LMM, resulting in enhanced multi-faceted explanations of IQA. Particularly, we propose a progressive instruction tuning strategy by dividing the adaption process of LMM for EIQA into two stages, where the first stage empowers the LMM with universal perception knowledge tailored for two tasks using an efficient transfer learning strategy, \ieno, LoRA, and the second stage introduces the instruction-adaptive visual prompt tuning to dynamically adapt visual features for the different instructions from two tasks. In this way, our proposed Q-Adapt can achieve a lightweight visual quality evaluator, demonstrating comparable performance and, in some instances, superior results across perceptual-related benchmarks and commonly-used IQA databases. The source code is publicly available at ~\url{https://github.com/yeppp27/Q-Adapt}.
\end{abstract}    
\vspace{-3mm}
\section{Introduction}
\label{sec:intro}
\vspace{-1mm}

Image Quality Assessment (IQA) aims to evaluate whether the image fidelity satisfies the human visual experience~\cite{qoe1,qoe2}, which has been used to various image processing techniques such as image compression~\cite{image_compression,image_compression2}, restoration~\cite{liang2021swinir, xia2023diffir}. However, despite that most IQA metrics, \egno, DEIQT~\cite{DEIQT}, LIPIPS~\cite{lpips} can provide an accurate quality score, they cannot explain the reasons in terms of distortions and contents behind the corresponding score. With the advancement of Large Multi-modal Foundation Models (LMM), Explainable Image Quality Assessment (EIQA) has become feasible due to the multi-modal reasoning and interaction capabilities of LMMs. A series of preliminary attempts have been made to excavate the low-level perception capability for images using LMMs~\cite{Q_bench,2AFC,Q_instruct}.

  \begin{figure}[htbp]
    \centering
    % 第一个图形
    \begin{minipage}[b]{0.2\textwidth}
        \includegraphics[width=\textwidth]{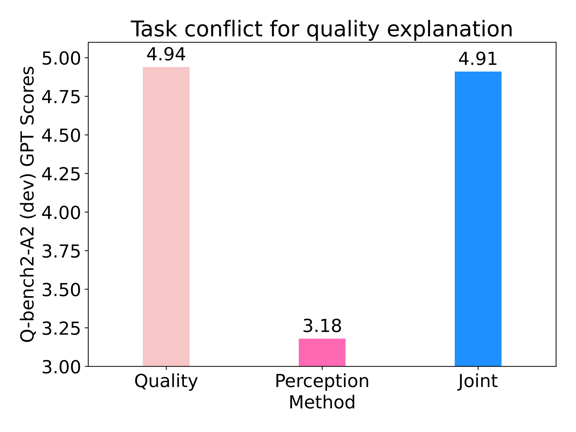}
        \caption{The effect of different task instruction tuning for quality explanation task.}
        \label{Co_taskconflict_cap}
    \end{minipage}
    \hfill % 添加空白填充以分隔相邻的 minipages
    % 第二个图形
    \begin{minipage}[b]{0.2\textwidth}
        \includegraphics[width=\textwidth]{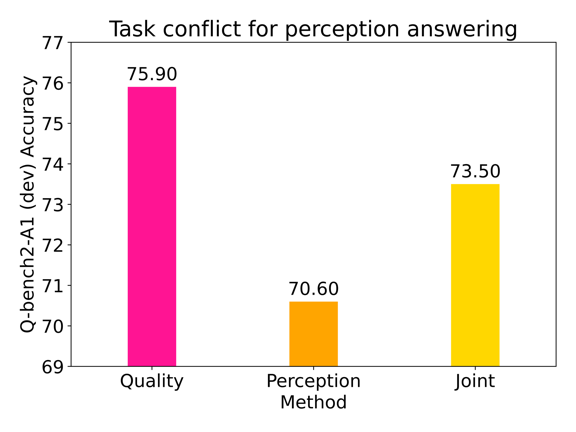}
        \caption{The effect of different task instruction tuning for perception answering task.}
        \label{Co_taskconflict_qa}
    \end{minipage}
    \hfill
    
\end{figure}

Existing works on LMM-based IQA can be roughly divided into two types. The first type aims to adapt the pre-trained LMMs to downstream IQA tasks by designing prompt templates, \ieno, prompt engineering, while freezing the parameters of LMMs. For instance, simply quality-aware prompt design can enable the GPT-4V~\cite{GPT-4V,Q_bench,Q_bench2} with great low-level visual perception capability. Despite the efficient adaptation, the frozen parameters limit the adequate low-level perception knowledge excavation required by downstream IQA tasks. The second type of works~\cite{Q_instruct,Co_instruct,depictQA} relies on instruction tuning, which aims to empower the pre-trained LMMs with overall quality explanation capability (\ieno, the left part of Fig.~\ref{fig:task}) and attribute-wise perception answering (\ieno, the right part of Fig.~\ref{fig:task}) capability by tuning the LMMs, preliminarily bridging the path to explainable IQA from two explanation perspectives.
From Fig.~\ref{Co_taskconflict_cap} and Fig.~\ref{Co_taskconflict_qa}, we observe that focusing exclusively on the explanation task improves performance compared to joint tuning of both tasks. Additionally, as illustrated in Fig.~\ref{fig:task}, Co-Instruct and GPT-4V exhibit instances of visual hallucinations in the question answering task. These observations highlight two fundamental challenges in LMM-based explainable image quality assessment:
(i) The conflicts between these two EIQA tasks are overlooked during instruction tuning, caused from the bias towards attribute-wise perception knowledge and the degradation of universal perception knowledge. (ii) The insufficient 
 cross-modal interaction restricts the adaptability to the synergy between these two EIQA tasks. As Fig.~\ref{fig:task} illustrates, insufficient reasoning capability and inflexible task instruction adaptation lead to misleading and spurious responses.

 To address the above issues, we propose Q-Adapt, a new paradigm for perception-oriented instruction tuning. Q-Adapt aims to eliminate task conflicts and achieve synergy between the two EIQA tasks, thereby enhancing the multifaceted explanations of IQA when adapting LMM as visual quality perceiver.
 Specifically, we propose a progressive instruction tuning by dividing the adaptation process of LMM for EIQA into two stages, continously enhancing perception knowledge for both tasks. The first stage involves the acquisition of universal perception knowledge in a parameter-efficient manner (\ieno, LoRA~\cite{hu2021lora}), establishing a powerful foundation that supports the different instruction requirements of both EIQA tasks.
 Building on the universal perception knowledge acquired in the first stage, we can more easily achieve adaptability for instructions across different tasks.
 However, the limited multimodal interactions~\cite {InternLM-XComposer2} within the layers of the LMM's language decoder are insufficient for adaptively capturing the visual knowledge specified by the instructions across both tasks.
To overcome this dilemma, we introduce instruction-adaptive visual prompt tuning, which dynamically adapts visual features to the different instructions, thereby enhancing the synergy between the two EIQA tasks.
In particular, to develop a visual prompt with powerful instruction adaptive capabilities, we employ bi-directional multimodal interactions to obtain an instruction-adaptive visual prompt, which consists of a vision-text (V-T) generator to fuse perception-related visual knowledge required by instructions into textual feature, and a text-vision (T-V) prompter that projects the textual feature back into the visual space.
The obtained instruction-adaptive visual prompt can guide the original visual feature through gated residual addition to highlight the crucial information specified by different instructions. Unlike uni-directional multimodal interactions (e.g., Q-Former~\cite{instructblip}), which capture condensed semantic information~\cite{yao2024deco} but lose fine-grained visual details, our bi-directional multimodal interaction module effectively acquires task-adaptive visual knowledge and refines the original visual feature without losing visual details.
In summary, the contributions of this paper are summarized as follows:
\begin{itemize}[leftmargin=*]
    \item We point out that simultaneously tuning LMMs with two types of Explainable Image quality Assessment (EIQA) tasks (\ieno, overall quality explanation and attribute-wise perception answering), can lead to potential task conflicts and insufficient perception understanding.
    \item To alleviate the above task conflicts, we introduce a new paradigm for perception-oriented instruction tuning, namely Q-Adapt. Q-Adapt employs a progressive instruction tuning which consists of two stages: the universal perception knowledge learning stage and the instruction-adaptive visual prompting stage. This approach achieves synergy between the two EIQA tasks and enhances the multifaceted explanations of IQA.
    \item Experimental results on perceptual-related benchmarks 
 and commonly-used IQA databases demonstrate that Q-Adapt achieves comparable and in some cases superior performance, even when utilizing a lightweight LMM model (\ieno, Bunny-3B~\cite{bunny}).

 \end{itemize}

 \section{Related Work}

\noindent \textbf{Large Multimodality Foundation Model}
 The large language models (LLM) have shown the powerful ability to act as a universal interface for a general-purpose assistant~\cite{zhang2023instruction}.
Following the step of LLM, LMMs are extended to conduct visual language tasks, which have achieved remarkable progress in multiple visual recognition and reasoning tasks~\cite{chen2023shikra, peng2023kosmos, ren2023pixellm, llava15}.
The cutting-edge works~\cite{llava15, Blip,instructblip} of LMM mainly bridge the visual encoder and LLM with a cross-modality connector to achieve the multimodal understanding ability. %BLIP2, Flagmingo, Llava
The milestone achievement, LLava~\cite{llava15} introduces visual instruction tuning to advance towards a general-purpose assistant.
And the following works in LMM can be divided into two categories: i) enhance visual perception, ii) enhance the interaction between visual and text representation.
For the first category, current works primarily optimize the visual representation by scaling the visual extractor or combining multiple visual experts.
From the perspective of the parameter scale of visual encoder, InternVL~\cite{internvl} scales up the visual encoder to match the parameter scale of LLM and proposes a progressive alignment strategy to harmonize the multimodal representations, which achieves outstanding ability in many vision-language tasks. Due to the limitation of CLIP visual encoder, Tong \textit{et. al}~\cite{tong2024eyes} interleaves the image feature from CLIP visual encoder and DINO~\cite{DINO,DINOv2} to enhance the visual grounding capabilities. Sphinx~\cite{sphinx} mixes image features from various visual encoders to achieve a versatile visual understanding ability. %Also, to enhance the fine-grained visual ability on region-level tasks, there are some works to encode the box or point as an additional prompt~\cite{} to fulfill the region-level visual understanding ability of a large language decoder.
%Besides, recent MLLMs are also extended to region-level parsing (Chen et al., 2023b; Peng et al., 2023b), in-context learning (Li et al., 2023a;b), arbitrary image resolutions (Bavishi et al., 2023), text-to-image generation (Wen et al., 2023; Dong et al., 2023), and 3D question answering ( from SPHINX
As for the second category, existing methods primarily focus on aligning visual and textual features before feeding into LLM or conducting visual-text collaboration/interaction within the deeper layers of the LLM.
To align visual features with task-specific instructions, InstructBLIP~\cite{instructblip} excavates the instruction-aware multimodal feature through Q-Former before integration into the LLM.
To implement multimodal collaboration, mPLUG-Owl2~\cite{mplug_owl2} processes visual and text features through different modules in each layer of LLM. With the same inspiration, CogVLM~\cite{Cogvlm} inserts the visual expert in each layer of LLM for deep alignment between two modalities. Inspired by the above two improvements, we aim to enhance the task-instruction adaptability of visual representation for multi-modal shallow alignment, thereby enabling the adaptive selection of the required granularity of perceptual knowledge.

\noindent \textbf{Large Multimodal Foundation Model for IQA.}
LMM for Image Quality Assessment (IQA) can be divided into three main streams. The first is to apply LMM to align the quality feature into text space. LIQE \cite{LIQE} fine-tuned the CLIP \cite{CLIP} model with fidelity loss to perceive the semantic-level scene, low-level distortion, and quality-level score. 
%ImageReward~\cite{Imagereward} fine-tune the BLIP~\cite{Blip} model on a self-built large scale AI-generated content image quality assessment dataset to capture the text-image alignment. 
Inspired by prompt learning for CLIP~\cite{Coop}, CLIPIQA~\cite{CLIPIQA} assesses quality scores by constructing prompt pairs with antonyms to evaluate the model's preference probability for score tokens. Through text generation, Q-Align~\cite{Q_align} enables LMM to evaluate quality scores that align with human opinions.
The second is using the prompt engineering technique to activate the quality perception ability of LMM. 
%QualiCLIP~\cite{QualiCLIP} employs a quality-aware alignment strategy to finetune CLIP, ranking degraded images by their similarity to quality-focused antonym text prompts.
Zhu \textit{et. al}~\cite{2AFC} employ two alternative forced choice (2AFC) prompting for multiple LMMs to explore their quality assessment ability.
To study more prompt strategy on LMM for quality assessment,
Wu \textit{et. al}~\cite{wu2024comprehensive} explores the chain-of-thought, in-context prompt to conduct the pair-wise image quality comparison. 
The third is to activate the instruction-following ability of LMM for explainable image quality assessment (EIQA). 
This line of research begins with the development of fine-grained low-level perceptual-related benchmark~\cite{AesBench, Q_bench, Q_bench2}, to evaluate the performance of both open-source~\cite{Minigpt-4, Internlm-xcomposer, mplug_owl2,llava15} and proprietary large multimodal models~\cite{GPT-4V,Gemini}. Subsequently, it involves the creation of the instruction datasets~\cite{Q_instruct, Co_instruct} that consists of the overall quality explanation task and attribute-wise perception answering task. These efforts aim to enhance the instruction-following ability of advanced multimodal large models for low-level vision. These approaches bridge the existing gap in IQA models regarding the capability for textual reasoning and interaction in an explainable manner. 
%Unlike these approaches, our method seeks to enable LMM to adapt to low-level perception through efficient training. 
In contrast to these approaches, our method facilitates the adaptation of LMMs to visual quality perception through efficient training. 
By mitigating the conflicts between the two EIQA tasks, we aim to achieve a more comprehensive understanding of visual quality perception.

\begin{figure*}[htp]
\centering
   \includegraphics[width=0.9\textwidth]{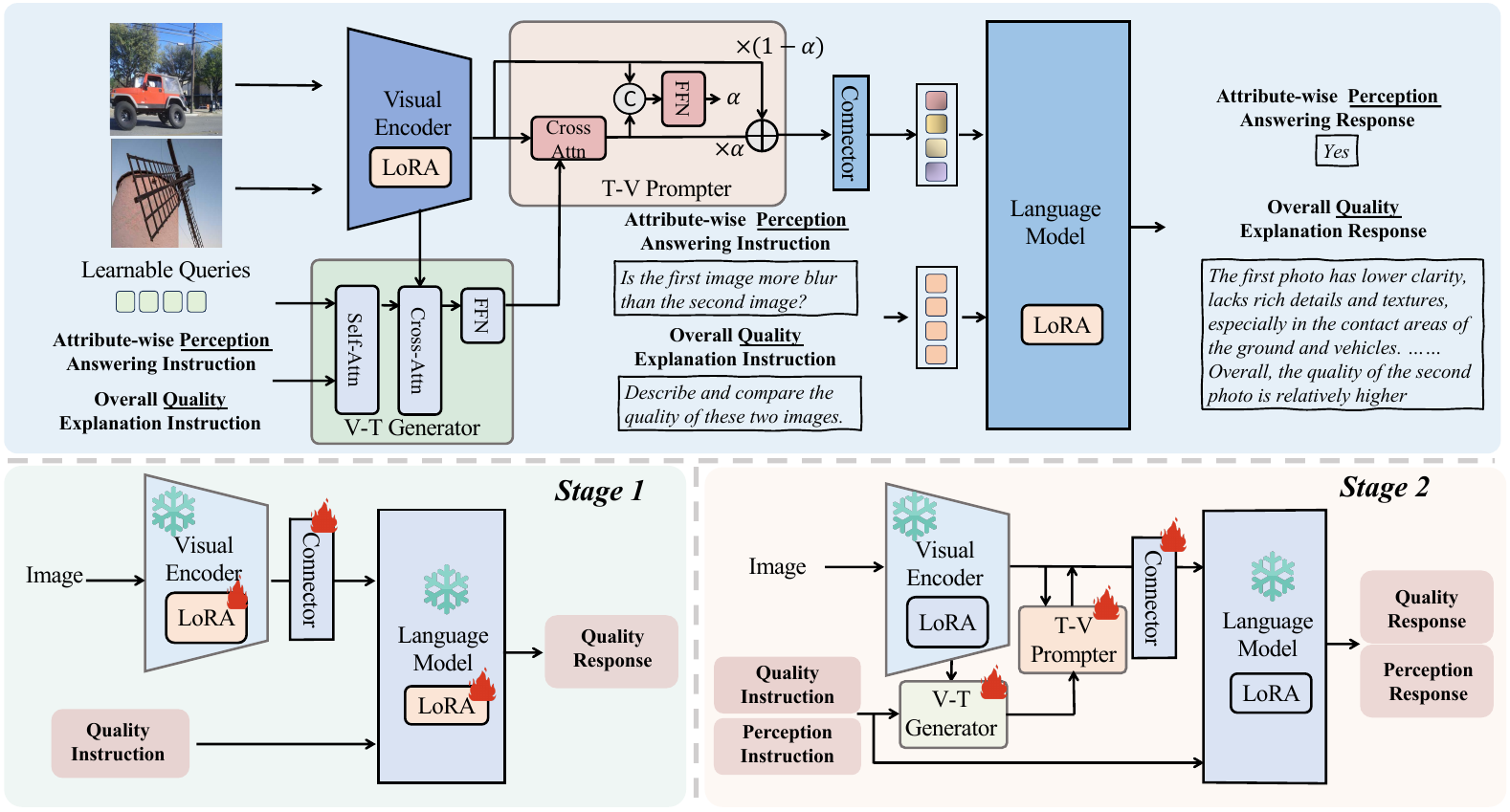}
    \vspace{-3mm}
    \caption{The overview for our proposed Q-Adapt, which employs progressive instruction tuning to achieve the synergy between two EIQA tasks. Concretely, the progressive instruction tuning strategy comprises two stages: the universal perception knowledge requiring stage (\ieno, the first stage) tailored for building a powerful base for two tasks, and the instruction-adaptive visual prompting stage for dynamically adapting visual features for task instruction. Additionally, the second stage incorporates the V-T Generator and T-V Prompter to achieve the bi-directional multimodal interactions.} %, which include video enhancement $\phi_{e}$, pre-processing $\phi_{p}$, and transcoding $\phi_{t}$.where $\phi_t$ applies uniformly to all videos using various QP values within $6$ bins, $\phi_e$ is triggered conditionally based on the predefined threshold, and $\phi_p$ randomly processes a subset proportion of original videos.   
    \label{fig:framework}
    \vspace{-3mm}
\end{figure*}
\vspace{-2mm}
\section{Method}
\vspace{-1mm}

\subsection{Preliminaries}
The primary objective of the Large Multi-modality Foundation Model (LMM) is to perceive visual signals and engage in reasoning through interactions with textual instructions, thereby addressing a variety of visual-language tasks.
The structure of the current LMM can be primarily summarized into three parts: the visual encoder, large language model (LLM), and multi-modal connector for bridging the visual and textual modality.

As for Explainable Image Quality Assessment (EIQA) task, given an image $v$ and perceptual-related instruction $I$, we extract the image feature $F_v \in \mathbb{R}^{n\times d_v}$ through the visual encoder, where $n$ is the number of visual tokens, and $d_v$ is the channel dimension. These features are subsequently processed through a connector $f_{vt}$, which maps them into the textual space, resulting in 
$F_{vt} \in \mathbb{R}^{n\times d_t}$, where $d_t$ represents the channel dimension, aligning with that of the text tokens. The transformed features, along with the instruction embedding $F_t\in \mathbb{R}^{m_t\times d_t}$, where $m_t$ denotes the number of the instruction tokens, are then fed into the Large Language Model (LLM). Optimization is performed using a language modeling loss based on next-token prediction~\cite{llava15,Llama}, which models the likelihood of the generated response conditioned on the provided images and instructions:
\vspace{-2mm}
\begin{equation}
L(r,v,I) = - \sum_{l=1}^{L_1} \log \left( P (r_l | v, I, r_{<l}) \right)
\end{equation}

Where $r_l$ represents the generated response token, conditioned on the input image $v$, instruction $I$, and previously generated response tokens $r_{<l}$.

\subsection{Task Conflicts for EIQA}
The Explainable Image Quality Assessment (EIQA) contain two tasks~\cite{Q_bench}: overall quality explanation~\cite{Q_instruct,depictQA}, and attribute-wise perception answering~\cite{Q_instruct}.
As shown in Fig.~\ref{fig:task},
The first task requires a long-text response detailing an overall quality explanation that integrates multiple low-level attributes and concludes with a final quality score.
The second task includes three types of perceptual-related visual question answering: multiple-choice, yes/no, and what/how questions, requiring brief answers for specific attributes/dimensions.

From Fig.~\ref{Co_taskconflict_qa}, we observe that tuning solely on the overall quality explanation task results in increased performance in the attribute-wise perception answering task, when compared to joint tuning on two tasks. 
It indicates that (i) an inherent conflict exists between the two tasks, since attribute-wise knowledge derived from training on the perception answering task tends to narrow the focus of the LMM towards localized/specific dimensions, lacking universal reasoning ability;
(ii ) the universal perception knowledge acquired through training on the quality explanation task explicitly assists in enhancing the reasoning capabilities for visual quality perception, which can build a powerful foundation.

\subsection{Progressive Instruction Tuning}

\subsubsection{Universal Perception Knowledge Learning Stage}

To address the conflicts between the two EIQA tasks,
we introduce the progressive instruction tuning strategy to enhance perception knowledge for the two EIQA tasks. 
It consists of two stages for perceptual-related instruction tuning on two tasks.
Based on the above observation, we are inspired to utilize the universal perception knowledge acquired from the overall quality explanation task to facilitate subsequent task adaption for different instructions.
Therefore, the first stage involves the instruction tuning on the quality explanation tasks for universal perception knowledge acquisition. To effectively learn the universal perception knowledge, this stage involves fine-tuning with a multimodal connector and utilizing the parameter-efficient LoRA~\cite{hu2021lora} technique on both the LLM and visual encoder.  
Specifically, the loss function of stage1 can be formulated as:
\vspace{-2mm}
\begin{equation}
L_{\text{stage1}}(a_{q},v,I_{q}) = - \sum_{l=1}^{L_1} \log \left( P_{\Phi_{0} + \Delta \Phi(\theta)} (a_{q,l} | v, I_{q}, a_{q,<l}) \right)
\end{equation}
where $\Phi_0$ and $\Delta \Phi(\theta)$ are referred to the parameters of frozen LMM and learnable LoRA parameters, respectively. And the subscript $q$ denotes the overall quality explanation task. $a_{q,l}$ represents the $l$-th token of the answer, and $I_{q}$ denotes the instruction of the overall quality explanation task. The $a_{q,<l}$ represents the generated answer token.
\vspace{-1mm}
\subsubsection{Instruction-guided Visual Prompt Tuning Stage}
In the second stage, to effectively enhance the perceptual knowledge for two EIQA tasks, two critical conditions must be fulfilled: (i) It is essential to adaptively select the required perception knowledge based on task instructions, which can alleviate the conflicts between the above two tasks. 
(ii) It is vital to ensure that the universal perception knowledge is not compromised by the attribute-wise knowledge from the attribute-wise perception answering task, thus enhancing the optimization of both tasks. 
Therefore, this stage requires fixing the parameters of the LLM and visual encoder, with the connector trainable, to prevent interference from biases towards specific perceptual knowledge for the single/localized dimension. 

Also, the self-attention mechanism in the LLM decoder treats visual and textual tokens equivalently across all layers~\cite{InternLM-XComposer2}, which limits its flexibility in extracting task-specific knowledge from visual features due to the insufficient cross-modal interactions.
Therefore, we propose the instruction-adaptive visual prompt tuning to excavate the essential knowledge required for the instruction for specific tasks. Concretely, we utilize the bidirectional interaction between instruction and visual features, which results in a prompt module comprising two specialized components: the V-T Generator, designed for vision-to-text interaction, and the T-V Prompter, tailored for text-to-vision interaction.

\noindent \textbf{V-T Generator}
 Due to the powerful vision-text interaction ability of the cross-attention-assisted transformer (\egno, Q-Former)~\cite{instructblip}, we leverage the Q-Former to enhance instruction representation with visual feature, enabling it to focus on informative visual knowledge for task instruction. 
Specifically, we input both the instruction representation $F_t$ and a fixed number of learnable queries $Q$ into the Q-Former.
%, utilizing cross-attention mechanisms to promote efficient vision-text interactions. 
This process yields an instruction representation $F_t$ that is enriched with visual features $F_v$, effectively bridging visual and textual representations and injecting the visual knowledge related to the instructions. The formulation of Q-Former is listed as follows:
\vspace{-2mm}
\begin{equation}
F_{vt} = \mathcal{G}(Q,F_t,f(F_v))
\end{equation}
Where, $Q\in \mathbb{R}^{m,d} $ denotes the learnable queries, $f(F_v)$ represents the projection for visual feature $F_v \in \mathbb{R}^{n,d_v}$ to match the dimension $d$. And the final obtained visual-guided instruction feature is $F_{vt}\in \mathbb{R}^{m,d}$. The V-T Generator (termed as $\mathcal{G}$), based on Q-Former (termed as $Q$), extracts instruction-adaptive visual features and maps them into the textual space, aggregating highly compressed perceptual information~\cite{yao2024deco} via a limited number of learnable queries, which results in a loss of fine-grained visual details. We then employ T-V Prompter to refine the original visual features, enabling the dynamic capture of task-related perceptual knowledge.

% Q-Former Process

\noindent \textbf{T-V Prompter}
To enhance the knowledge adaptation of the original visual features, we introduce a second stage of text-vision interaction. As depicted in Fig.~\ref{fig:framework}, this stage employs a gated fusion process to generate an instruction-adaptive visual prompt. Specifically, we utilize cross-attention to integrate the information from highly-condensed 
 multimodal feature $F_{vt}$ into the original visual feature $F_v$, facilitating the dynamic modulation of the original visual feature. Subsequently, a sigmoid-gated fusion mechanism is applied to merge the intermediate feature $\tilde{F}_{tv} \in {\mathbb{R}^{n,d_v}}$ with the original visual feature $F_v \in \mathbb{R}^{n,d_v}$.
\vspace{-1mm}
 \begin{align}  
\tilde{F}_{tv} &= \text{CA}(F_v, f(F_{vt}), f(F_{vt})) \\
F_{tv} &= (1 - \sigma(\tilde{F}_{tv},F_v)) \tilde{F}_{tv} + \sigma(\tilde{F}_{tv}, F_v) F_v
\end{align}

Where $f(\cdot)$ is utilized to map the channel dimension $d$ of $F_{vt}$ to $d_v$. CA denotes the cross attention mechanism between $F_v$ and $f(F_{vt})$. And $\sigma(\cdot)$ computes the weights for gated fusion.
Through the above operations, we can modulate the original visual features through the gated residual addition, effectively integrating the instruction-adaptive visual prompt to refine the original visual feature.
%Therefore, the final loss for the second stage can be updated:
%\vspace{-2mm}
%\begin{align}
%L_{\text{stage2}}(a, v, I) &= 
%- \sum_{l=1}^{L_1} \log \left( P_{\Phi_2} (a_{q,l} | v, I_{q,<l}, a_{q,<l}) \right) \\
%&\quad - \sum_{l=1}^{L_2} \log \left( P_{\Phi_2} (a_{a,l} | v, I_{a,<l}, a_{a,<l}) \right)
%\end{align}
%where, $\Phi_2 =\Phi_1 + \Theta_{p}$, $\Theta_{p}$ is denoted as the parameters of our learnable prompt modules. And the subscript $q$ denotes the overall quality explanation task and $a$ denotes the attribute-wise question answering task.

\begin{figure*}[htbp]
   \centering
\includegraphics[width=1\textwidth]{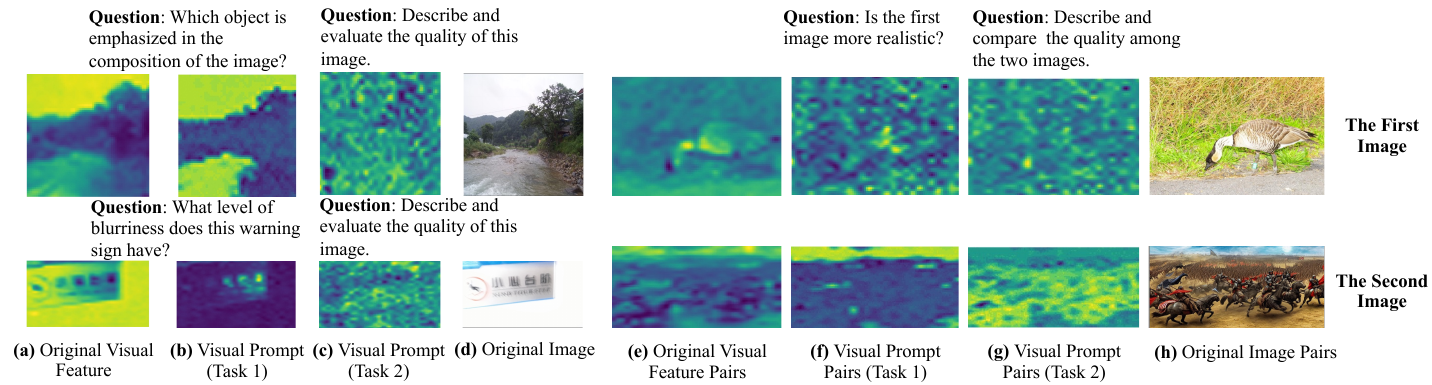}
    \vspace{-5mm}
    \caption{The visualizations of the original visual feature and instruction-adaptive visual prompt. (a)-(d) illustrate results from Q-bench, while (e)-(h) show results from Q-bench2. And ``Task1" refers to the attribute-wise perception answering task, ``Task2" denotes the overall quality explanation task.} 
    \label{fig:visualization}
\end{figure*}

 \begin{table*}[htbp]
\centering
\footnotesize
\caption{Comparison of Different Methods for visual question answering task. }
\vspace{-2mm}
\label{tab:comparison}
\resizebox{0.83\textwidth}{!}{\begin{tabular}{l|ccc|ccc| ccc }
\toprule
\textbf{Method}  & \multicolumn{3}{c|}{\textbf{Q-bench-A1 (\%)}}  & \multicolumn{3}{|c}{\textbf{Q-bench2-A1 (\%)}}  & \multicolumn{3}{|c}{\textbf{MME }}  \\ %\cline{2-5} 
                &              \textbf{dev} & \textbf{test}  & \textbf{Average}  & \textbf{dev} & \textbf{test} & \textbf{Average}    & \textbf{Perception} & \textbf{Cognition}  & \textbf{Score}                              \\ \midrule
Bunny-3B~\cite{bunny}\textit{(Baseline)}            & 65.08           &   64.68  & 64.88     &48.20         &50.85   &     49.53      & 1488  &  289   & 1777  \\ 
LLaVA-v1.5-13B~\cite{llava15}        & 62.14            & 61.40     &  61.77    & 49.85         & 52.05  &     50.95      & 1476  &   313 &  1743         \\ 
mPLUG-Owl2~\cite{mplug_owl2}                  & 61.61            & 62.68     &   62.15      & 49.85        & 48.94 &49.40    & 1450  &  313               & 1763      \\  
Emu2-Chat~\cite{emu2}                 & 65.28            & 64.32    &      64.80   & 50.05         & 47.08    & 48.57     & -  & -    & -                   \\ 
Qwen2-VL~\cite{wang2024qwen2} &78.13 &- &- &75.6 &- &-  & -&- & 2326\\
LLaVA-NeXT-Interleave-7B~\cite{li2024llava} & 73.58&- &- &74.2 & -&- &- &- &1778 \\
LLaVA-OneVision~\cite{onevision} &77.66 & -&- &76.5 &- &-  &1580& 418&1998\\
\midrule

Qwen-VL-Max~\cite{bai2023qwen}            &73.63            &73.90     &    73.77   & 67.27          & 66.99   &  67.13    & -  &-      & 2281    \\ 
Gemini-Pro~\cite{Gemini}          & 68.16            & 69.46  &  68.81     & 57.64            &60.46         & 59.02      & -  &  -         & -    \\ 
GPT-4V ~\cite{GPT-4V}        & 74.51            & 74.10       & 74.31  & 76.52            &78.07        &     77.30       &  1409 &   517          &   1926        \\ \midrule

Co-Instruct-8B~\cite{Co_instruct}                   & 76.99           & 77.12   &  77.05   & 78.40       & 80.18  &   79.29    & 1266  & 303   &  1569                \\ 
\rowcolor{gray!30}Q-Adapt-3B$^{Co}$               & 76.05           & 76.12   & 76.08    & 77.20      & 78.38     &  77.79   &  1313  &  286  & 1599\\ \midrule

Q-Instruct-8B~\cite{Q_instruct}                &70.23           & 73.38 &  71.81   & 50.54       & 53.15   & 51.85   &1443  & 337 &1780  \\    
%& 4.04             & 3.69              \\ 
\rowcolor{gray!30}Q-Adapt-3B$^{Q}$                & 77.19         & 77.06   &   77.12    &  55.40      &  55.96     &  55.68   &  1343 &   271   &  1614         
\\\bottomrule
% Repeat for other rows as necessary
\end{tabular}}
\vspace{-5mm}
\label{exp_1}
\end{table*}

\begin{table*}[htbp]
\centering
\caption{Performance comparison on overall quality explanation task. We employ the 5-round GPT score as defined in~\cite{Q_bench2} for our evaluation metric.  Here, $P_i$ denotes the frequency of a rating in the set of 0, 1 and 2. A higher GPT score indicates better performance.}
\resizebox{\textwidth}{!}{\begin{tabular}{l|cccc|cccc|cccc|c}
\hline
\textbf{Dimensions} & \multicolumn{4}{c|}{\textbf{Completeness}} & \multicolumn{4}{c|}{\textbf{Precision}} & \multicolumn{4}{c|}{\textbf{Relevance}} & \multirow{2}{*}{\textbf{Sum}} \\
\textbf{Model} & \multicolumn{1}{c}{$P_0$} & $P_1$ & $P_2$ & \textbf{score} & \multicolumn{1}{c}{$P_0$} & $P_1$ & $P_2$ & \textbf{score} & \multicolumn{1}{c}{$P_0$} & $P_1$ & $P_2$ & \textbf{score} & \\
\hline
Bunny-3B~\cite{bunny}  & 24.40\% & 71.64\% & 3.95\% & 0.79 & 9.86\% & 50.53\% & 39.60\% & 1.29  & 0.97\% & 21.73\% & 77.28\% & 1.76 & 3.85 \\
LLaVA-v1.5-13B~\cite{llava15} & 18.77\% & 73.44\% & 7.79\% & 0.89 & 34.66\% & 38.72\% & 26.62\% & 0.92 & 1.02\% & 34.59\% & 64.39\% & 1.63 & 3.44 \\
mPLUG-Owl2~\cite{mplug_owl2} & 19.43\% & 65.54\% & 14.45\% &0.94 & 30.94\% & 43.71\% & 24.63\% & 0.92 & 3.79\% & 26.94\% & 68.28\% & 1.63 & 3.50 \\
Emu2-Chat~\cite{emu2} 
&41.25\%	&54.33\%	&4.42\%	 &0.63	&38.11\%	&36.41\%	&25.48\%	&0.87	&4.12\%	&38.61\%	&57.27\%	&1.53	&3.03 \\
%& 41.25\% & 54.38\% & 4.42\% & 0.90 & 38.11\% & 36.41\% & 25.48\% & 0.94 & 4.12\% & 38.61\% & 57.27\% & 2.73 & 4.56 \\
Qwen-VL-Max~\cite{bai2023qwen}
&11.64\%	&54.08\%	&34.08\%	&1.22	&24.26\%	&39.14\%	&36.22\%	&1.11	&2.53\%	&10.97\%	&85.64\%	&1.82	&4.16 \\

Gemini-Pro~\cite{Gemini} & 18.22\% & 44.48\% & 36.84\% & 1.18 & 34.13\% & 37.95\% & 27.02\% & 0.92 & 0.67\% & 5.91\% & 92.22\% & 1.90 & 4.00 \\
GPT-4V~\cite{GPT-4V} & 4.09\% & 31.82\% & 64.09\% & 1.60 & 10.40\% & 45.12\% & 44.44\% & 1.34 & 0.18\% & 1.69\% & 96.35\% & 1.94 & 4.89 \\

Co-Instruct~\cite{Co_instruct} & 4.04\% & 31.55\% & 63.55\% & 1.58 & 13.68\% & 43.68\% & 41.37\% & 1.26 & 0.0\% & 0.44\% & 98.22\% & 1.96 & 4.82 \\
%Q-Adapt$^{co}$ & 18.56\% & 26.34\% &55.10\% &  1.37 & \% & \% & \% & 1.66 & 0.03\% & 0.44\% & 95.59\% & 1.95 & 4.98 \\
\rowcolor{gray!30}Q-Adapt$^{co}$ &8.97\%  & 44.22\%&46.79\% &1.38  &3.82\% & 27.15\% & 69.02\% &  1.65& 0.0\% & 4.17\% & 95.8\% & 1.96 &4.98\\
\hline
\end{tabular}}
\label{exp_2}
\end{table*}

\begin{table*}[htbp]
\centering

\caption{The comparison results of quality assessment (SROCC/PLCC).}
\resizebox{\textwidth}{!}{\begin{tabular}{@{}lcccccccc@{}}
\toprule
\textbf{Model} & \textbf{KonIQ-10k} & \textbf{SPAQ} & \textbf{LIVE-FB} & \textbf{LIVE-itw}  & \textbf{AGIQA-3k}
& \textbf{CGIQA-6k} & \textbf{KADID-10k} & \textbf{Average}  \\ \midrule
LIQE~\cite{LIQE}  &0.897/0.914& 0.925/0.922 & 0.469/0.541& 0.868/0.884 &0.744/0.807& 0.161/0.197& 0.675/0.663& 0.677/0.704\\
LoDa~\cite{LoDa} &0.804/0.844 &0.892/0.899 &0.460/0.524& 0.784/0.820 &0.687/0.744 &0.303/0.322   &0.636/0.649&0.653/0.686  \\
LLaVA-v1.5~\cite{llava15}  &0.448/0.460& 0.563/0.584& 0.310/0.339& 0.445/0.481 &0.285/0.297& 0.664/0.754& 0.390/0.400& 0.444/0.474 \\
mPLUG-Owl2~\cite{mplug_owl2}&0.196/0.252 &0.589/0.614 &0.217/0.286& 0.293/0.342 &0.473/0.492&-0.024/-0.032  &0.541/0.546& 0.326/0.357 \\
Emu2-Chat~\cite{emu2} &0.664/0.714 &0.712/0.698 &0.355/0.341 &0.597/0.611  &0.759/0.751&0.224/0.269& 0.841/0.790 &0.593/0.596 \\
InternLM-XComposer-VL~\cite{Internlm-xcomposer} &0.564/0.615& 0.730/0.750 &0.360/0.416 &0.612/0.676 &0.732/0.775 &0.243/0.265 &0.546/0.572& 0.541/0.581 \\ \midrule
Co-Instruct~\cite{Co_instruct} &0.839/0.898  & 0.869/0.900 &\textbf{0.467/0.584  } &0.839/0.851  &0.680/0.708& 0.421/0.438 &0.762/0.756   & 0.696/0.733  \\ 
\rowcolor{gray!30}Q-Adapt$^{Co}$ & \textbf{0.869/0.898}& \textbf{0.916/0.915}  & 0.460/0.539  & \textbf{0.869/0.897} & \textbf{0.739/0.783}  &  0.429/0.435 & 0.720/0.711  & \textbf{0.714/0.739}  \\ \midrule
Q-Instruct~\cite{Q_instruct}             & \textbf{0.911/0.921}         & 0.901/0.898          & \textbf{0.442/0.535}            &   0.842/0.840   &  0.700/0.763      &  0.572/0.578  &  0.682/0.683 &0.721/0.745\\

\rowcolor{gray!30}Q-Adapt$^{Q}$ & 0.878/0.907 & \textbf{0.913/0.916} & 0.440/0.517 & 0.837/\textbf{0.845}  &  \textbf{0.757/0.789} & \textbf{0.593/0.595}  & \textbf{0.769/0.754} &\textbf{ 0.741/0.760}   \\  
 
\bottomrule
\end{tabular}}
\label{exp_3}
\end{table*}
\section{Experiment}

\subsection{Datasets and Implementation Details}

\label{sec:Datasets and Implementation Details}
\noindent \textbf{Training Datasets} We conduct the perceptual-oriented visual instruction tuning on two datasets: Q-Instruct~\cite{Q_instruct} and Co-Instruct~\cite{Co_instruct}. Q-Instruct has a total of 200k instruction-response pairs. Besides, Co-Instruct extends Q-Instruct from single image to multiple images, which includes 580k instruction-response pairs. The model trained on Q-Instruct and Co-Instruct is named Q-Adapt$^{Q}$, Q-Adapt$^{Co}$.

\noindent \textbf{Evaluation Benchmarks} We evaluate our proposed Q-Adapt on the challenging perceptual-related benchmark Q-bench-A1~\cite{Q_bench} and Q-bench2-A1~\cite{Q_bench2} for the attribute-wise perception answering task, and Q-bench2-A2~\cite{Q_bench2} for the overall quality explanation task. We also select commonly-used benchmark MME~\cite{fu2023mme} for high-level task evaluation.
We also tested the performance of our Q-Adapt on commonly-used IQA datasets for quality assessment~\cite{koniq,spaq,flive,livec,agiqa3k,cgiiqa,kadid}.

\subsection{Comparison Results}
To verify the effectiveness of our proposed method, we evaluate our proposed Q-Adapt against two types of Large Multi-modal Foundation Models (LMMs): a frozen-based LMM and an instruction-tuning-based LMM. Some of  frozen-based models (\egno, GPT-4V~\cite{GPT-4V}, Gemini-pro~\cite{Gemini} and Qwen-max~\cite{Gemini}) are proprietary and closed-source. 
The performance of most of these frozen-based LMMs is generally inferior as they have not been exposed to image-quality-related textual data during previous training. Notably, within these comparative methods, \textbf{our Q-Adapt employs a parameter-efficient tuning strategy, and the total parameter size is only 3B.} 

\noindent \textbf{Attribute-wise Perception Answering Task.}
The results of performance comparison on the perception answering task are shown in Table~\ref{exp_1}.
For Q-bench-A1, Q-Adapt$^Q$ surpasses the second-best method, Q-Instruct-8B, by a margin of 5.31\% on average accuracy. And our Q-Adapt$^{Co}$, with a parameter size of 3B and LoRA training, achieves performance close to Co-instruct-8B on Q-bench2-A1.

\noindent \textbf{Overall Quality Explanation Task.}
For Q-bench2-A2, the comparison results are represented in Table~\ref{exp_2}. Our Q-Adapt$^{Co}$ achieves a performance gain of 0.09 over the second-best method GPT-4V on the GPT score. It is attributed to our ability to achieve synergy between the two EIQA tasks, thereby improving perception precision. More examples can be found in \textbf{Appendix}.

\noindent \textbf{Image Quality Assessment.} We also evaluate the performance of Q-Adapt$^{Q}$ on multiple IQA databases and compare it with existing LMMs and IQA models. For IQA models, LIQE~\cite{LIQE} and LoDa~\cite{LoDa} utilize networks to regress predicted scores against quality annotations. We transform the Q-Instruct dataset from image-text pairs to image-score pairs to facilitate regression for both LoDa and LIQE. From Table~\ref{exp_3}, Q-Adapt$^{Q}$ can achieve the best performance compared to other methods on the average performance of SROCC/PLCC. It is noteworthy that our Q-Adapt significantly outperforms existing LMMs and quality assessment models on the AGIQA-3k~\cite{agiqa3k}, CGIQA-6k~\cite{cgiiqa}, and KADID-10k~\cite{kadid} datasets, which are barely existed in the training process. It underscores the strong generalization ability of Q-Adapt, which can be attributed to the parameter-efficient training approach.

\begin{table}[htbp]
%\begin{wraptable}{r}{0.5\textwidth}
\centering
\footnotesize
\vspace{-4mm}
\caption{Parameters and FLOPs comparisons for different models, with performance metrics computed on the Q-bench-A1-dev.}
\vspace{-2mm}
\resizebox{0.4\textwidth}{!}{\begin{tabular}{@{}lccc@{}}
\toprule
\textbf{} &\textbf{Q-Instruct-8B } & \textbf{Bunny-3B (LoRA)} & \textbf{Q-Adapt-3B }  \\ \midrule
%Latency      & 347.27ms      & 334.49ms       & 365.23ms           \\
Flops & 1700G &  656.18 G & 695.32 G     \\ %\midrule
Param &8.2B & \ 2.78B & 2.98B    \\ \midrule
Performance &70.23            &  69.57        & \textbf{ 77.19}          \\

\bottomrule
\end{tabular}}
\vspace{-2mm}
\label{exp_flops_latency}
\end{table}

\noindent \textbf{Parameters and Flops.} Q-Adapt presents an effective tuning strategy that utilizes minimal parameter increases to achieve substantial performance improvements over the baseline model, Bunny-3B, as well as the more parameter-intensive Q-Instruct-8B, thereby offering a more efficient solution for EIQA task adaptation from the well-built LMM.
%theoretical
%\FloatBarrier

\begin{table}[htbp]
\centering
\footnotesize
\caption{Ablation study for instruction-guided visual prompt.}
\resizebox{0.44\textwidth}{!}{\begin{tabular}{@{}lccc|ccc@{}}
\toprule
\textbf{} & \textbf{Q-bench-A1 (dev)} & \textbf{Q-bench-A1 (test)} &\textbf{Average} & \textbf{Q-bench2-A1 (dev)} & \textbf{Q-bench-A2 (test)}&\textbf{Average} \\ \midrule
w.o. prompt$^{Q}$            & 74.45       & 75.25      & 74.85    &  52.50          &  51.85    &52.17       \\
Q-Adapt$^{Q}$ & \textbf{ 77.19 } & \textbf{77.06 }  & \textbf{77.12} &   \textbf{55.40}  &   \textbf{55.96} & \textbf{55.68}    \\ \midrule
w.o. prompt$^{Co}$             &  75.93        & \textbf{ 75.71}   &75.82      &   76.80          &  76.77    &76.78       \\
Q-Adapt$^{Co}$ & \textbf{76.05} & 76.12 &  \textbf{76.08} & \textbf{77.20}   & \textbf{78.38}    & \textbf{77.79}\\ 
\bottomrule
\end{tabular}}
\label{exp_prompt1}
\end{table}

\vspace{-5mm}
\subsection{Ablation Study}
%\paragraph{The Effectiveness of Variants of Instruction-guided Visual Prompt.} 

%The first approach focuses on the encoder structure for generating prompts, as shown in Table~\ref{exp_prompt2}. The second approach involves the multimodal interaction techniques to be employed during prompt generation, as presented in Table~\ref{exp_prompt3}. The third approach examines the position and manner in which the prompts modulate the original visual features, as detailed in Table~\ref{exp_prompt3}.

\noindent (I) \textbf{The Existence of Instruction-guided Visual Prompt.} The effectiveness of instruction-guided visual prompt for Q-Adapt in the Stage 2 training phase is explored in Table~\ref{exp_prompt1}. In the Table, "w.o. prompt" indicates that only the multimodal connector is trainable. From the results, it is evident that with the assistance of the instruction-guided visual prompt, Q-Adapt achieves a performance gain over training only the connector. It highlights the effect of the instruction-guided visual prompt in adaptively excavating perceptual knowledge required by task instructions.
As shown in Fig.~\ref{fig:visualization}, we demonstrate the effectiveness of our proposed instruction-adaptive visual prompting. 
The visualization results indicate that, for the question answering task, the instruction-adaptive features concentrate on areas specified by the instruction or corresponding to potential answers. In contrast, the  visual prompt for the overall quality explanation task typically highlights a broader range of visual details. This demonstrates a dynamic modulation for two EIQA tasks.

%More ablation study for instruction-adaptive visual prompting can be found in \textbf{Appendix}~\ref{sec:app_Instruction-Adaptive Visual Prompting}

\begin{table*}[hthp]
\centering
\footnotesize
\caption{Ablation study on progressive instruction tuning on Q-Instruct dataset.}
\resizebox{\textwidth}{!}{\begin{tabular}{@{}c|cc|cccc|ccc@{}}
\toprule
 \multicolumn{1}{c|}{\textbf{Training Stages}}& \multicolumn{2}{c|}{\textbf{Tasks}}
&\multicolumn{4}{c|}{\textbf{Module}} &\multicolumn{3}{c}{\textbf{ Q-bench}}\\
\multicolumn{1}{c|}{}& \textbf{Quality} &\multicolumn{1}{c|}{\textbf{Perception}} & \textbf{Vision LoRA} & \textbf{LLM LoRA} & \textbf{Connector} & \multicolumn{1}{c|}{\textbf{Prompt Module}} & \textbf{dev} & \textbf{test} & \textbf{Average} \\
\midrule
\multirow{4}{*}{Stage 1}  &  {\ding{51}}  &{\ding{55}}  &{\ding{51}} &{\ding{51}} &{\ding{51}} &{\ding{55}} &\textbf{73.51} &\textbf{73.31} & \textbf{73.41}\\

    &  {\ding{55}}  &{\ding{51}}  &{\ding{51}} &{\ding{51}} &{\ding{51}} &{\ding{55}} & 67.96 & 69.83 &68.89 \\
    
    &  {\ding{51}}  &{\ding{51}}  &{\ding{51}} &{\ding{51}} &{\ding{51}} &{\ding{55}} & 69.57 & 69.89& 69.73\\ 
    &  {\ding{51}}  &{\ding{51}}  &{\ding{51}} &{\ding{51}} &{\ding{51}} &{\ding{51}} &71.30 & 74.38 & 72.84\\ \midrule
  \multirow{5}{*}{Stage 2} &  {\ding{51}}  &{\ding{51}}  &{\ding{55}} &{\ding{55}} &{\ding{51}} &{\ding{51}}&\textbf{77.19} &\textbf{77.06} & \textbf{77.12}\\
  & {\ding{51}}  &{\ding{55}}  &{\ding{55}} &{\ding{55}} &{\ding{51}} &{\ding{51}}&70.10 &69.40 & 69.75\\
   &  {\ding{55}}  &{\ding{51}}  &{\ding{55}} &{\ding{55}} &{\ding{51}} &{\ding{51}} & 75.59& 75.45 &75.52\\
   &  {\ding{51}}  &{\ding{51}}  &{\ding{55}} &{\ding{55}} &{\ding{55}} &{\ding{51}} & 74.85&74.11&74.48 \\
   &  {\ding{51}}  &{\ding{51}}  &{\ding{55}} &{\ding{51}} &{\ding{51}} &{\ding{51}} & 74.45& 75.98&75.21\\
\bottomrule
\end{tabular}}
\label{exp_progressive}
\end{table*}

%\paragraph{The Effectiveness of Dual Instruction Tuning. }

\begin{figure} % l 表示左侧环绕, 0.35\textwidth 表示图片宽度为文本宽度的一半
\vspace{-2mm}
    \centering
    \includegraphics[width=0.4\textwidth]{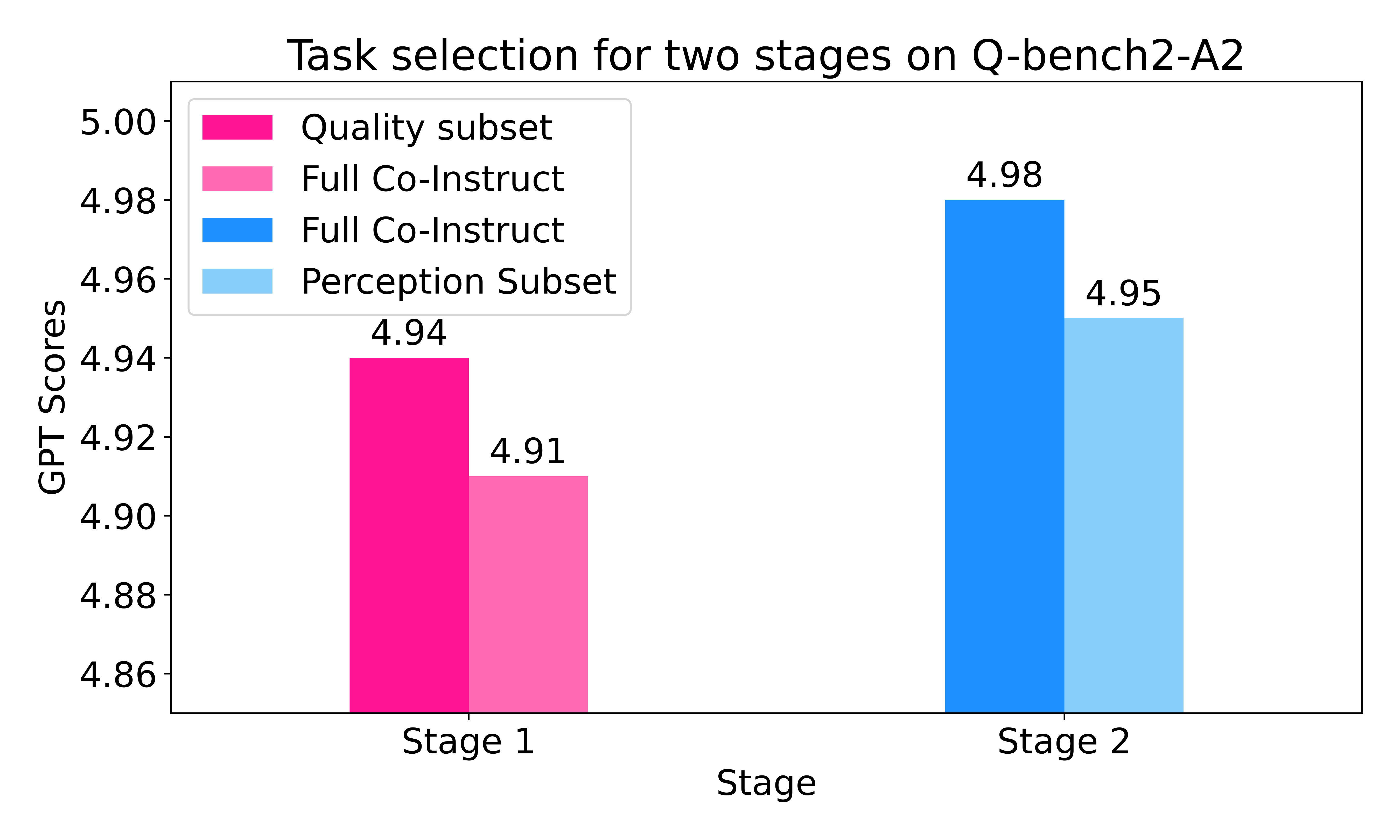} % 替换为您的图片文件
    \vspace{-6mm}
    \caption{The effect of task selection in progressive instruction tuning for explanation task.}
    \vspace{-6mm}
     \label{fig:caption_ablation}
\end{figure}

\noindent (II)\textbf{The Effectiveness of Progressive Instruction Tuning.}
We analyze the effect of progressive instruction tuning for training on Q-Instruct in Table~\ref{exp_progressive}. Additionally, we examine the impact of task selection for overall quality explanation tasks, as shown in Fig.~\ref{fig:caption_ablation}. And we also conduct a comprehensive comparison across different models in Table~\ref{exp_different_backbone_tuning} for joint tuning on two EIQA tasks, two-stage tuning, and our proposed progressive-instruction tuning. 
%More ablation study for progressive instruction tuning can be found in \textbf{Appendix}~\ref{sec:app_Dual Instruction Tuning}.
\textit{\textbf{The Task for Instruction Tuning.}} For the first stage of instruction tuning  (\ieno, universal perception knowledge learning stage), the results (the $1^{st}$, $2^{nd}$, and $3^{rd}$ rows of Table~\ref{exp_progressive}) show that the performance of joint tuning on both tasks and only tuning on the perception answering task are lower than tuning on the overall quality explanation task. Also, from Fig.~\ref{fig:caption_ablation}, we can see that the performance can be boosted when training on the Quality subset (\ieno, overall quality explanation). It reflects the inherent conflicts between the two tasks. 
For the second stage of instruction tuning (\ieno, the instruction-adaptive visual prompting stage),  the results ($5^{th}$ and $6^{th}$ rows of Table~\ref{exp_progressive}) demonstrate that joint tuning for both tasks yields an average accuracy gain of 1.6\% compared to tuning exclusively on the perception answering task. The similar phenomenon is observed in the quality explanation task in Fig.~\ref{fig:caption_ablation},  removing the explanation subset results in a performance decline (from 4.98 to 4.95). 
%\tcr{Additionally, training solely on the quality explanation task in Stage 2 leads to a significant performance decline. This is due to the excessive focus on universal global reasoning, which compromises the model's ability to effectively address question answering tasks.} It underscores the significance of achieving synergy between the two EIQA tasks in the second stage. 
\textit{\textbf{The Trainable Modules.}}
For the first stage tuning (\ieno, the universal perception knowledge learning stage), the results for trainable modules are shown in the $3^{rd}$ and $4^{th}$ rows of Table~\ref{exp_progressive}. The findings reveal that joint tuning on the prompt module results in an average accuracy improvement of 3.11\%, demonstrating the effectiveness of instruction-adaptive visual prompts for adapting to different instructions. However, it is still lower than only training on quality explanation tasks, due to the importance of required universal perception knowledge.
The results in the second stage (\ieno, the instruction-adaptive visual prompting stage) is examined in the $5^{th}$, $7^{th}$, and $8^{th}$ rows of Table~\ref{exp_progressive}. We draw two conclusions from the results: Firstly, a trainable multimodal connector is essential for the second stage of instruction tuning, since it plays a critical role in modality alignment. Secondly, a trainable LoRA for the language decoder is unnecessary in the second stage, as the language decoder should remain fixed to preserve the universal perceptual knowledge acquired in the first stage. 

\begin{table}[htbp]%[!h]
\centering
\caption{The comparisons between different models and tuning strategies on Q-bench-A1 (dev), where all methods utilize LoRA for efficient training. ``Pro. Ins. Tuning" denotes ``progressive instruction tuning".} 
\vspace{-3mm}
\resizebox{0.45\textwidth}{!}{\begin{tabular}{@{}lccc@{}}
\toprule
& \textbf{Joint Tuning} & \textbf{Two-Stage Tuning}  & \textbf{Pro. Ins. Tuning}  \\ \midrule
LLama-VID-8B~\cite{Llama-vid}            & 65.55        &  63.81    &67.49     \\
mPLUG-Owl2-8B~\cite{mplug_owl2}   & 66.69 & 67.76  & 69.03   \\ \midrule

Bunny-3B~\cite{bunny}          &  69.57  &  68.28      & \textbf{ 77.19}          \\

\bottomrule
\end{tabular}}
\vspace{-4mm}
\label{exp_different_backbone_tuning}
\end{table}

\noindent (III) \textbf{Progressive Instruction Tuning across different backbones.} We present a comprehensive comparison of LLama-VID~\cite{Llama-vid}, mPLUG-Owl2~\cite{mplug_owl2}, and Bunny~\cite{bunny} across joint tuning, two-stage tuning, and our proposed progressive instruction tuning on Q-Instruct dataset, as detailed in Table~\ref{exp_different_backbone_tuning}. All training strategies utilize LoRA for efficient training.
The two-stage tuning approach consists of two phases: initially training on the overall quality explanation task with a trainable multimodal connector for alignment, followed by training on the two EIQA tasks using both the connector and the LLM. Experimental results in the table indicate that progressive instruction tuning yields the best performance, as it effectively mitigates task conflict. In contrast, the two-stage tuning process, which resembles the training strategy of existing LMMs, is inadequate for adapting LMMs to downstream tasks, such as EIQA. More ablation studies can be found in our \textbf{Supplementary Materials}.

\section{Conclusion}

In summary, to alleviate the inherent conflicts in two EIQA tasks (\ieno, overall quality explanation, and attribute-wise perception answering), we propose Q-Adapt to adapt LMM as a visual quality perceiver, which is conducted through a perception-oriented instruction tuning strategy, namely, progressive instruction tuning. The progressive instruction tuning consists of the universal perception learning stage for building a powerful base for two tasks, and the instruction-adaptive prompting stage for dynamically adapting visual features for different instructions. By doing this, our Q-Adapt can achieve the synergy between these two EIQA tasks when adapting LMM. Extension experiments on two related benchmarks can illustrate the effectiveness of our Q-Adapt on both overall quality explanation task and attribute-wise perception answering task. 
%\input{sec/4_conclusion}
%input{sec/4_conclusion}
%\input{sec/2_formatting}
%\input{sec/3_finalcopy}
{
    \small
    \bibliographystyle{ieeenat_fullname}
    \bibliography{main}

\begin{thebibliography}{61}
\providecommand{\natexlab}[1]{#1}
\providecommand{\url}[1]{\texttt{#1}}
\expandafter\ifx\csname urlstyle\endcsname\relax
  \providecommand{\doi}[1]{doi: #1}\else
  \providecommand{\doi}{doi: \begingroup \urlstyle{rm}\Url}\fi

\bibitem[Bai et~al.(2023)Bai, Bai, Yang, Wang, Tan, Wang, Lin, Zhou, and Zhou]{bai2023qwen}
Jinze Bai, Shuai Bai, Shusheng Yang, Shijie Wang, Sinan Tan, Peng Wang, Junyang Lin, Chang Zhou, and Jingren Zhou.
\newblock Qwen-vl: A versatile vision-language model for understanding, localization, text reading, and beyond.
\newblock \emph{arXiv preprint arXiv:2308.12966}, 2023.

\bibitem[Caron et~al.(2021)Caron, Touvron, Misra, J{\'e}gou, Mairal, Bojanowski, and Joulin]{DINO}
Mathilde Caron, Hugo Touvron, Ishan Misra, Herv{\'e} J{\'e}gou, Julien Mairal, Piotr Bojanowski, and Armand Joulin.
\newblock Emerging properties in self-supervised vision transformers.
\newblock In \emph{Proceedings of the IEEE/CVF international conference on computer vision}, pages 9650--9660, 2021.

\bibitem[Chen et~al.(2023{\natexlab{a}})Chen, Zhang, Zeng, Zhang, Zhu, and Zhao]{chen2023shikra}
Keqin Chen, Zhao Zhang, Weili Zeng, Richong Zhang, Feng Zhu, and Rui Zhao.
\newblock Shikra: Unleashing multimodal llm's referential dialogue magic.
\newblock \emph{arXiv preprint arXiv:2306.15195}, 2023{\natexlab{a}}.

\bibitem[Chen et~al.(2023{\natexlab{b}})Chen, Wu, Wang, Su, Chen, Xing, Muyan, Zhang, Zhu, Lu, et~al.]{internvl}
Zhe Chen, Jiannan Wu, Wenhai Wang, Weijie Su, Guo Chen, Sen Xing, Zhong Muyan, Qinglong Zhang, Xizhou Zhu, Lewei Lu, et~al.
\newblock Internvl: Scaling up vision foundation models and aligning for generic visual-linguistic tasks.
\newblock \emph{arXiv preprint arXiv:2312.14238}, 2023{\natexlab{b}}.

\bibitem[Dai et~al.(2024)Dai, Li, Li, Tiong, Zhao, Wang, Li, Fung, and Hoi]{instructblip}
Wenliang Dai, Junnan Li, Dongxu Li, Anthony Meng~Huat Tiong, Junqi Zhao, Weisheng Wang, Boyang Li, Pascale~N Fung, and Steven Hoi.
\newblock Instructblip: Towards general-purpose vision-language models with instruction tuning.
\newblock \emph{Advances in Neural Information Processing Systems}, 36, 2024.

\bibitem[Darcet et~al.(2023)Darcet, Oquab, Mairal, and Bojanowski]{DINOv2}
Timoth{\'e}e Darcet, Maxime Oquab, Julien Mairal, and Piotr Bojanowski.
\newblock Vision transformers need registers.
\newblock \emph{arXiv preprint arXiv:2309.16588}, 2023.

\bibitem[Devlin et~al.(2018)Devlin, Chang, Lee, and Toutanova]{Bert}
Jacob Devlin, Ming-Wei Chang, Kenton Lee, and Kristina Toutanova.
\newblock Bert: Pre-training of deep bidirectional transformers for language understanding.
\newblock \emph{arXiv preprint arXiv:1810.04805}, 2018.

\bibitem[Dong et~al.(2024)Dong, Zhang, Zang, Cao, Wang, Ouyang, Wei, Zhang, Duan, Cao, et~al.]{InternLM-XComposer2}
Xiaoyi Dong, Pan Zhang, Yuhang Zang, Yuhang Cao, Bin Wang, Linke Ouyang, Xilin Wei, Songyang Zhang, Haodong Duan, Maosong Cao, et~al.
\newblock Internlm-xcomposer2: Mastering free-form text-image composition and comprehension in vision-language large model.
\newblock \emph{arXiv preprint arXiv:2401.16420}, 2024.

\bibitem[Fang et~al.(2020)Fang, Zhu, Zeng, Ma, and Wang]{spaq}
Yuming Fang, Hanwei Zhu, Yan Zeng, Kede Ma, and Zhou Wang.
\newblock Perceptual quality assessment of smartphone photography.
\newblock In \emph{Proceedings of the IEEE/CVF conference on computer vision and pattern recognition}, pages 3677--3686, 2020.

\bibitem[Fu et~al.(2023)Fu, Chen, Shen, Qin, Zhang, Lin, Yang, Zheng, Li, Sun, et~al.]{fu2023mme}
Chaoyou Fu, Peixian Chen, Yunhang Shen, Yulei Qin, Mengdan Zhang, Xu Lin, Jinrui Yang, Xiawu Zheng, Ke Li, Xing Sun, et~al.
\newblock Mme: A comprehensive evaluation benchmark for multimodal large language models.
\newblock \emph{arXiv preprint arXiv:2306.13394}, 2023.

\bibitem[Ghadiyaram and Bovik(2015)]{livec}
Deepti Ghadiyaram and Alan~C Bovik.
\newblock Massive online crowdsourced study of subjective and objective picture quality.
\newblock \emph{IEEE Transactions on Image Processing}, 25\penalty0 (1):\penalty0 372--387, 2015.

\bibitem[Google(2023)]{Gemini}
Google.
\newblock Gemini pro.
\newblock \emph{https://deepmind. google/technologies/gemini}, 2023.

\bibitem[He et~al.(2024)He, Liu, Wu, Yuan, Wang, Huang, and Zhao]{bunny}
Muyang He, Yexin Liu, Boya Wu, Jianhao Yuan, Yueze Wang, Tiejun Huang, and Bo Zhao.
\newblock Efficient multimodal learning from data-centric perspective.
\newblock \emph{arXiv preprint arXiv:2402.11530}, 2024.

\bibitem[Hosu et~al.(2020)Hosu, Lin, Sziranyi, and Saupe]{koniq}
Vlad Hosu, Hanhe Lin, Tamas Sziranyi, and Dietmar Saupe.
\newblock Koniq-10k: An ecologically valid database for deep learning of blind image quality assessment.
\newblock \emph{IEEE Transactions on Image Processing}, 29:\penalty0 4041--4056, 2020.

\bibitem[Hu et~al.(2021)Hu, Shen, Wallis, Allen-Zhu, Li, Wang, Wang, and Chen]{hu2021lora}
Edward~J Hu, Yelong Shen, Phillip Wallis, Zeyuan Allen-Zhu, Yuanzhi Li, Shean Wang, Lu Wang, and Weizhu Chen.
\newblock Lora: Low-rank adaptation of large language models.
\newblock \emph{arXiv preprint arXiv:2106.09685}, 2021.

\bibitem[Huang et~al.(2024)Huang, Yuan, Sheng, Yang, Wu, Chen, Yang, Li, and Lin]{AesBench}
Yipo Huang, Quan Yuan, Xiangfei Sheng, Zhichao Yang, Haoning Wu, Pengfei Chen, Yuzhe Yang, Leida Li, and Weisi Lin.
\newblock Aesbench: An expert benchmark for multimodal large language models on image aesthetics perception.
\newblock \emph{arXiv preprint arXiv:2401.08276}, 2024.

\bibitem[Li et~al.(2024{\natexlab{a}})Li, Zhang, Guo, Zhang, Li, Zhang, Zhang, Zhang, Li, Liu, et~al.]{onevision}
Bo Li, Yuanhan Zhang, Dong Guo, Renrui Zhang, Feng Li, Hao Zhang, Kaichen Zhang, Peiyuan Zhang, Yanwei Li, Ziwei Liu, et~al.
\newblock Llava-onevision: Easy visual task transfer.
\newblock \emph{arXiv preprint arXiv:2408.03326}, 2024{\natexlab{a}}.

\bibitem[Li et~al.(2023{\natexlab{a}})Li, Zhang, Wu, Sun, Min, Liu, Zhai, and Lin]{agiqa3k}
Chunyi Li, Zicheng Zhang, Haoning Wu, Wei Sun, Xiongkuo Min, Xiaohong Liu, Guangtao Zhai, and Weisi Lin.
\newblock Agiqa-3k: An open database for ai-generated image quality assessment.
\newblock \emph{IEEE Transactions on Circuits and Systems for Video Technology}, 2023{\natexlab{a}}.

\bibitem[Li et~al.(2024{\natexlab{b}})Li, Zhang, Zhang, Zhang, Li, Li, Ma, and Li]{li2024llava}
Feng Li, Renrui Zhang, Hao Zhang, Yuanhan Zhang, Bo Li, Wei Li, Zejun Ma, and Chunyuan Li.
\newblock Llava-next-interleave: Tackling multi-image, video, and 3d in large multimodal models.
\newblock \emph{arXiv preprint arXiv:2407.07895}, 2024{\natexlab{b}}.

\bibitem[Li et~al.(2022)Li, Li, Xiong, and Hoi]{Blip}
Junnan Li, Dongxu Li, Caiming Xiong, and Steven Hoi.
\newblock Blip: Bootstrapping language-image pre-training for unified vision-language understanding and generation.
\newblock In \emph{International conference on machine learning}, pages 12888--12900. PMLR, 2022.

\bibitem[Li et~al.(2023{\natexlab{b}})Li, Wang, and Jia]{Llama-vid}
Yanwei Li, Chengyao Wang, and Jiaya Jia.
\newblock Llama-vid: An image is worth 2 tokens in large language models.
\newblock \emph{arXiv preprint arXiv:2311.17043}, 2023{\natexlab{b}}.

\bibitem[Liang et~al.(2021)Liang, Cao, Sun, Zhang, Van~Gool, and Timofte]{liang2021swinir}
Jingyun Liang, Jiezhang Cao, Guolei Sun, Kai Zhang, Luc Van~Gool, and Radu Timofte.
\newblock Swinir: Image restoration using swin transformer.
\newblock In \emph{Proceedings of the IEEE/CVF international conference on computer vision}, pages 1833--1844, 2021.

\bibitem[Lin et~al.(2019)Lin, Hosu, and Saupe]{kadid}
Hanhe Lin, Vlad Hosu, and Dietmar Saupe.
\newblock Kadid-10k: A large-scale artificially distorted iqa database.
\newblock In \emph{2019 Eleventh International Conference on Quality of Multimedia Experience (QoMEX)}, pages 1--3. IEEE, 2019.

\bibitem[Lin et~al.(2023)Lin, Liu, Zhang, Gao, Qiu, Xiao, Qiu, Lin, Shao, Chen, et~al.]{sphinx}
Ziyi Lin, Chris Liu, Renrui Zhang, Peng Gao, Longtian Qiu, Han Xiao, Han Qiu, Chen Lin, Wenqi Shao, Keqin Chen, et~al.
\newblock Sphinx: The joint mixing of weights, tasks, and visual embeddings for multi-modal large language models.
\newblock \emph{arXiv preprint arXiv:2311.07575}, 2023.

\bibitem[Liu et~al.(2023)Liu, Li, Li, and Lee]{llava15}
Haotian Liu, Chunyuan Li, Yuheng Li, and Yong~Jae Lee.
\newblock Improved baselines with visual instruction tuning.
\newblock \emph{arXiv preprint arXiv:2310.03744}, 2023.

\bibitem[Moller et~al.(2009)Moller, Engelbrecht, Kuhnel, Wechsung, and Weiss]{qoe1}
Sebastian Moller, Klaus-Peter Engelbrecht, Christine Kuhnel, Ina Wechsung, and Benjamin Weiss.
\newblock A taxonomy of quality of service and quality of experience of multimodal human-machine interaction.
\newblock In \emph{2009 international workshop on quality of multimedia experience}, pages 7--12. IEEE, 2009.

\bibitem[OpenAI(2023)]{GPT-4V}
OpenAI.
\newblock Gpt-4v(ision) system card.
\newblock \emph{https://cdn.openai.com/papers/ GPTV System Card.pdf/}, 2023.

\bibitem[Peng et~al.(2023)Peng, Wang, Dong, Hao, Huang, Ma, and Wei]{peng2023kosmos}
Zhiliang Peng, Wenhui Wang, Li Dong, Yaru Hao, Shaohan Huang, Shuming Ma, and Furu Wei.
\newblock Kosmos-2: Grounding multimodal large language models to the world.
\newblock \emph{arXiv preprint arXiv:2306.14824}, 2023.

\bibitem[Qin et~al.(2023)Qin, Hu, Liu, Zheng, Liu, Li, and Zhang]{DEIQT}
Guanyi Qin, Runze Hu, Yutao Liu, Xiawu Zheng, Haotian Liu, Xiu Li, and Yan Zhang.
\newblock Data-efficient image quality assessment with attention-panel decoder.
\newblock In \emph{Proceedings of the AAAI Conference on Artificial Intelligence}, pages 2091--2100, 2023.

\bibitem[Radford et~al.(2021)Radford, Kim, Hallacy, Ramesh, Goh, Agarwal, Sastry, Askell, Mishkin, Clark, et~al.]{CLIP}
Alec Radford, Jong~Wook Kim, Chris Hallacy, Aditya Ramesh, Gabriel Goh, Sandhini Agarwal, Girish Sastry, Amanda Askell, Pamela Mishkin, Jack Clark, et~al.
\newblock Learning transferable visual models from natural language supervision.
\newblock In \emph{International conference on machine learning}, pages 8748--8763. PMLR, 2021.

\bibitem[Reiter et~al.(2014)Reiter, Brunnstr{\"o}m, De~Moor, Larabi, Pereira, Pinheiro, You, and Zgank]{qoe2}
Ulrich Reiter, Kjell Brunnstr{\"o}m, Katrien De~Moor, Mohamed-Chaker Larabi, Manuela Pereira, Antonio Pinheiro, Junyong You, and Andrej Zgank.
\newblock Factors influencing quality of experience.
\newblock \emph{Quality of experience: Advanced concepts, applications and methods}, pages 55--72, 2014.

\bibitem[Ren et~al.(2023)Ren, Huang, Wei, Zhao, Fu, Feng, and Jin]{ren2023pixellm}
Zhongwei Ren, Zhicheng Huang, Yunchao Wei, Yao Zhao, Dongmei Fu, Jiashi Feng, and Xiaojie Jin.
\newblock Pixellm: Pixel reasoning with large multimodal model.
\newblock \emph{arXiv preprint arXiv:2312.02228}, 2023.

\bibitem[Sun et~al.(2023)Sun, Cui, Zhang, Zhang, Yu, Luo, Wang, Rao, Liu, Huang, et~al.]{emu2}
Quan Sun, Yufeng Cui, Xiaosong Zhang, Fan Zhang, Qiying Yu, Zhengxiong Luo, Yueze Wang, Yongming Rao, Jingjing Liu, Tiejun Huang, et~al.
\newblock Generative multimodal models are in-context learners.
\newblock \emph{arXiv preprint arXiv:2312.13286}, 2023.

\bibitem[Tong et~al.(2024)Tong, Liu, Zhai, Ma, LeCun, and Xie]{tong2024eyes}
Shengbang Tong, Zhuang Liu, Yuexiang Zhai, Yi Ma, Yann LeCun, and Saining Xie.
\newblock Eyes wide shut? exploring the visual shortcomings of multimodal llms.
\newblock \emph{arXiv preprint arXiv:2401.06209}, 2024.

\bibitem[Touvron et~al.(2023)Touvron, Lavril, Izacard, Martinet, Lachaux, Lacroix, Rozi{\`e}re, Goyal, Hambro, Azhar, et~al.]{Llama}
Hugo Touvron, Thibaut Lavril, Gautier Izacard, Xavier Martinet, Marie-Anne Lachaux, Timoth{\'e}e Lacroix, Baptiste Rozi{\`e}re, Naman Goyal, Eric Hambro, Faisal Azhar, et~al.
\newblock Llama: Open and efficient foundation language models.
\newblock \emph{arXiv preprint arXiv:2302.13971}, 2023.

\bibitem[Wang et~al.(2025)Wang, Cui, Li, Lin, Chen, and Zhang]{VTC}
Dongsheng Wang, Jiequan Cui, Miaoge Li, Wang Lin, Bo Chen, and Hanwang Zhang.
\newblock Instruction tuning-free visual token complement for multimodal llms.
\newblock In \emph{European Conference on Computer Vision}, pages 446--462. Springer, 2025.

\bibitem[Wang et~al.(2023{\natexlab{a}})Wang, Chan, and Loy]{CLIPIQA}
Jianyi Wang, Kelvin~CK Chan, and Chen~Change Loy.
\newblock Exploring clip for assessing the look and feel of images.
\newblock In \emph{Proceedings of the AAAI Conference on Artificial Intelligence}, pages 2555--2563, 2023{\natexlab{a}}.

\bibitem[Wang et~al.(2024)Wang, Bai, Tan, Wang, Fan, Bai, Chen, Liu, Wang, Ge, et~al.]{wang2024qwen2}
Peng Wang, Shuai Bai, Sinan Tan, Shijie Wang, Zhihao Fan, Jinze Bai, Keqin Chen, Xuejing Liu, Jialin Wang, Wenbin Ge, et~al.
\newblock Qwen2-vl: Enhancing vision-language model's perception of the world at any resolution.
\newblock \emph{arXiv preprint arXiv:2409.12191}, 2024.

\bibitem[Wang et~al.(2023{\natexlab{b}})Wang, Lv, Yu, Hong, Qi, Wang, Ji, Yang, Zhao, Song, et~al.]{Cogvlm}
Weihan Wang, Qingsong Lv, Wenmeng Yu, Wenyi Hong, Ji Qi, Yan Wang, Junhui Ji, Zhuoyi Yang, Lei Zhao, Xixuan Song, et~al.
\newblock Cogvlm: Visual expert for pretrained language models.
\newblock \emph{arXiv preprint arXiv:2311.03079}, 2023{\natexlab{b}}.

\bibitem[Wu et~al.(2023{\natexlab{a}})Wu, Zhang, Zhang, Chen, Liao, Wang, Li, Sun, Yan, Zhai, et~al.]{Q_bench}
Haoning Wu, Zicheng Zhang, Erli Zhang, Chaofeng Chen, Liang Liao, Annan Wang, Chunyi Li, Wenxiu Sun, Qiong Yan, Guangtao Zhai, et~al.
\newblock Q-bench: A benchmark for general-purpose foundation models on low-level vision.
\newblock \emph{arXiv preprint arXiv:2309.14181}, 2023{\natexlab{a}}.

\bibitem[Wu et~al.(2023{\natexlab{b}})Wu, Zhang, Zhang, Chen, Liao, Wang, Xu, Li, Hou, Zhai, et~al.]{Q_instruct}
Haoning Wu, Zicheng Zhang, Erli Zhang, Chaofeng Chen, Liang Liao, Annan Wang, Kaixin Xu, Chunyi Li, Jingwen Hou, Guangtao Zhai, et~al.
\newblock Q-instruct: Improving low-level visual abilities for multi-modality foundation models.
\newblock \emph{arXiv preprint arXiv:2311.06783}, 2023{\natexlab{b}}.

\bibitem[Wu et~al.(2023{\natexlab{c}})Wu, Zhang, Zhang, Chen, Liao, Li, Gao, Wang, Zhang, Sun, et~al.]{Q_align}
Haoning Wu, Zicheng Zhang, Weixia Zhang, Chaofeng Chen, Liang Liao, Chunyi Li, Yixuan Gao, Annan Wang, Erli Zhang, Wenxiu Sun, et~al.
\newblock Q-align: Teaching lmms for visual scoring via discrete text-defined levels.
\newblock \emph{arXiv preprint arXiv:2312.17090}, 2023{\natexlab{c}}.

\bibitem[Wu et~al.(2024{\natexlab{a}})Wu, Zhu, Zhang, Zhang, Chen, Liao, Li, Wang, Sun, Yan, et~al.]{Co_instruct}
Haoning Wu, Hanwei Zhu, Zicheng Zhang, Erli Zhang, Chaofeng Chen, Liang Liao, Chunyi Li, Annan Wang, Wenxiu Sun, Qiong Yan, et~al.
\newblock Towards open-ended visual quality comparison.
\newblock \emph{arXiv preprint arXiv:2402.16641}, 2024{\natexlab{a}}.

\bibitem[Wu et~al.(2024{\natexlab{b}})Wu, Ma, Liang, Yang, and Zhang]{wu2024comprehensive}
Tianhe Wu, Kede Ma, Jie Liang, Yujiu Yang, and Lei Zhang.
\newblock A comprehensive study of multimodal large language models for image quality assessment.
\newblock \emph{arXiv preprint arXiv:2403.10854}, 2024{\natexlab{b}}.

\bibitem[Wu et~al.(2021)Wu, Li, Zhang, Jin, and Chen]{image_compression2}
Yaojun Wu, Xin Li, Zhizheng Zhang, Xin Jin, and Zhibo Chen.
\newblock Learned block-based hybrid image compression.
\newblock \emph{IEEE Transactions on Circuits and Systems for Video Technology}, 32\penalty0 (6):\penalty0 3978--3990, 2021.

\bibitem[Xia et~al.(2023)Xia, Zhang, Wang, Wang, Wu, Tian, Yang, and Van~Gool]{xia2023diffir}
Bin Xia, Yulun Zhang, Shiyin Wang, Yitong Wang, Xinglong Wu, Yapeng Tian, Wenming Yang, and Luc Van~Gool.
\newblock Diffir: Efficient diffusion model for image restoration.
\newblock In \emph{Proceedings of the IEEE/CVF International Conference on Computer Vision}, pages 13095--13105, 2023.

\bibitem[Xu et~al.(2024)Xu, Liao, Xiao, Chen, Wu, Yan, and Lin]{LoDa}
Kangmin Xu, Liang Liao, Jing Xiao, Chaofeng Chen, Haoning Wu, Qiong Yan, and Weisi Lin.
\newblock Boosting image quality assessment through efficient transformer adaptation with local feature enhancement.
\newblock In \emph{Proceedings of the IEEE/CVF Conference on Computer Vision and Pattern Recognition}, pages 2662--2672, 2024.

\bibitem[Yang and Mandt(2024)]{image_compression}
Ruihan Yang and Stephan Mandt.
\newblock Lossy image compression with conditional diffusion models.
\newblock \emph{Advances in Neural Information Processing Systems}, 36, 2024.

\bibitem[Yao et~al.(2024)Yao, Li, Ren, Wang, Liu, Sun, and Hou]{yao2024deco}
Linli Yao, Lei Li, Shuhuai Ren, Lean Wang, Yuanxin Liu, Xu Sun, and Lu Hou.
\newblock Deco: Decoupling token compression from semantic abstraction in multimodal large language models.
\newblock \emph{arXiv preprint arXiv:2405.20985}, 2024.

\bibitem[Ye et~al.(2023)Ye, Xu, Ye, Yan, Liu, Qian, Zhang, Huang, and Zhou]{mplug_owl2}
Qinghao Ye, Haiyang Xu, Jiabo Ye, Ming Yan, Haowei Liu, Qi Qian, Ji Zhang, Fei Huang, and Jingren Zhou.
\newblock mplug-owl2: Revolutionizing multi-modal large language model with modality collaboration.
\newblock \emph{arXiv preprint arXiv:2311.04257}, 2023.

\bibitem[Ying et~al.(2020)Ying, Niu, Gupta, Mahajan, Ghadiyaram, and Bovik]{flive}
Zhenqiang Ying, Haoran Niu, Praful Gupta, Dhruv Mahajan, Deepti Ghadiyaram, and Alan Bovik.
\newblock From patches to pictures (paq-2-piq): Mapping the perceptual space of picture quality.
\newblock In \emph{Proceedings of the IEEE/CVF conference on computer vision and pattern recognition}, pages 3575--3585, 2020.

\bibitem[You et~al.(2023)You, Li, Gu, Yin, Xue, and Dong]{depictQA}
Zhiyuan You, Zheyuan Li, Jinjin Gu, Zhenfei Yin, Tianfan Xue, and Chao Dong.
\newblock Depicting beyond scores: Advancing image quality assessment through multi-modal language models.
\newblock \emph{arXiv preprint arXiv:2312.08962}, 2023.

\bibitem[Zhang et~al.(2023{\natexlab{a}})Zhang, Wang, Cao, Xu, Ouyang, Zhao, Ding, Zhang, Duan, Yan, et~al.]{Internlm-xcomposer}
Pan Zhang, Xiaoyi Dong~Bin Wang, Yuhang Cao, Chao Xu, Linke Ouyang, Zhiyuan Zhao, Shuangrui Ding, Songyang Zhang, Haodong Duan, Hang Yan, et~al.
\newblock Internlm-xcomposer: A vision-language large model for advanced text-image comprehension and composition.
\newblock \emph{arXiv preprint arXiv:2309.15112}, 2023{\natexlab{a}}.

\bibitem[Zhang et~al.(2018)Zhang, Isola, Efros, Shechtman, and Wang]{lpips}
Richard Zhang, Phillip Isola, Alexei~A Efros, Eli Shechtman, and Oliver Wang.
\newblock The unreasonable effectiveness of deep features as a perceptual metric.
\newblock In \emph{Proceedings of the IEEE conference on computer vision and pattern recognition}, pages 586--595, 2018.

\bibitem[Zhang et~al.(2023{\natexlab{b}})Zhang, Dong, Li, Zhang, Sun, Wang, Li, Hu, Zhang, Wu, et~al.]{zhang2023instruction}
Shengyu Zhang, Linfeng Dong, Xiaoya Li, Sen Zhang, Xiaofei Sun, Shuhe Wang, Jiwei Li, Runyi Hu, Tianwei Zhang, Fei Wu, et~al.
\newblock Instruction tuning for large language models: A survey.
\newblock \emph{arXiv preprint arXiv:2308.10792}, 2023{\natexlab{b}}.

\bibitem[Zhang et~al.(2023{\natexlab{c}})Zhang, Zhai, Wei, Yang, and Ma]{LIQE}
Weixia Zhang, Guangtao Zhai, Ying Wei, Xiaokang Yang, and Kede Ma.
\newblock Blind image quality assessment via vision-language correspondence: A multitask learning perspective.
\newblock In \emph{Proceedings of the IEEE/CVF conference on computer vision and pattern recognition}, pages 14071--14081, 2023{\natexlab{c}}.

\bibitem[Zhang et~al.(2023{\natexlab{d}})Zhang, Sun, Zhou, Jia, Zhang, Liu, Min, and Zhai]{cgiiqa}
Zicheng Zhang, Wei Sun, Yingjie Zhou, Jun Jia, Zhichao Zhang, Jing Liu, Xiongkuo Min, and Guangtao Zhai.
\newblock Subjective and objective quality assessment for in-the-wild computer graphics images.
\newblock \emph{ACM Transactions on Multimedia Computing, Communications and Applications}, 20\penalty0 (4):\penalty0 1--22, 2023{\natexlab{d}}.

\bibitem[Zhang et~al.(2024)Zhang, Wu, Zhang, Zhai, and Lin]{Q_bench2}
Zicheng Zhang, Haoning Wu, Erli Zhang, Guangtao Zhai, and Weisi Lin.
\newblock A benchmark for multi-modal foundation models on low-level vision: from single images to pairs.
\newblock \emph{arXiv preprint arXiv:2402.07116}, 2024.

\bibitem[Zhou et~al.(2022)Zhou, Yang, Loy, and Liu]{Coop}
Kaiyang Zhou, Jingkang Yang, Chen~Change Loy, and Ziwei Liu.
\newblock Learning to prompt for vision-language models.
\newblock \emph{International Journal of Computer Vision}, 130\penalty0 (9):\penalty0 2337--2348, 2022.

\bibitem[Zhu et~al.(2023)Zhu, Chen, Shen, Li, and Elhoseiny]{Minigpt-4}
Deyao Zhu, Jun Chen, Xiaoqian Shen, Xiang Li, and Mohamed Elhoseiny.
\newblock Minigpt-4: Enhancing vision-language understanding with advanced large language models.
\newblock \emph{arXiv preprint arXiv:2304.10592}, 2023.

\bibitem[Zhu et~al.(2024)Zhu, Sui, Chen, Liu, Chen, Fang, and Wang]{2AFC}
Hanwei Zhu, Xiangjie Sui, Baoliang Chen, Xuelin Liu, Peilin Chen, Yuming Fang, and Shiqi Wang.
\newblock 2afc prompting of large multimodal models for image quality assessment.
\newblock \emph{arXiv preprint arXiv:2402.01162}, 2024.

\end{thebibliography}
}

\section*{Appendix}
\section{Experiment Details}
\noindent \textbf{Implementation Details}
%We use Bunny-3B~\citep{} as the baseline for our proposed method. 
To construct the V-T Generator module, the Q-Former module in
InstructBLIP~\citep{instructblip} is applied as our V-T Generator.
The number of queries in V-T Generator is 32, which follows previous work. And the cross-attention in T-V Prompter only has a single head.
Given that LMM is often constrained by their substantial computational costs and model parameters, we have adopted Bunny-3B~\citep{bunny}, one of the lightweight multimodal model families for instruction tuning. The training of Q-Adapt requires two 32G V100 GPUs for training, and one 32G V100 GPU for testing.
 %More details can be found in the \textbf{Appendix}~\ref{sec: app_Implementation Details}.
 
\noindent \textbf{The Encoder Structure for V-T Generator.} The analysis for the encoder strure of V-T Generator is shown in Table~\ref{exp_prompt2}. Utilizing the Q-Former~\citep{instructblip} can achieve an average accuracy increase of 1.43\% on Q-bench-A1 for instruction tuning on Q-Instruct, compared to the BERT~\citep{Bert} structure. It demonstrates that the Q-Former, by introducing learnable queries, can capture high-level semantic information from the instructions, facilitating the extraction of crucial task information.

\noindent \textbf{Training Details.}
 %We train the Q-Adapt with AdamW optimizer, and the
%earning rate is set as $2e^{-5}$. 
The detailed of hyperparameters and modules are listed below:
Visual Encoder: siglip-so400m-patch14-384, LLM: phi-2, image resolution: 384, batchsize: 64, learning rate: 2e-5, learning rate schedule: cosine decay, weight decay: 0, warmup ratio: 0.03, gradient accumulation steps: 4, numerical precision: float16, epochs for stage 1: 1, epochs for stage 2: 1, optimizer: AdamW, deepspeed stage2.

Following the pioneering works of LMM paradigm~\citep{Q_instruct} of finetuning strategy and model architecture, we inherit weights from the Bunny-3B of instruction version to apply continual instruction tuning to downstream EIQA tasks. In the progressive instruction tuning approach applied to the Q-Instruct dataset, the first stage solely focuses on the overall quality explanation task to acquire universal knowledge. The second stage involves joint tuning across the full Q-Instruct dataset. For the Co-Instruct dataset, given that the baseline model, Bunny-3B, has not been exposed to multiple images for vision question answering, we transform the attribute-wise perception answering task data into chain-of-thought quality data (i.e., multi-turn conversations). This data is then combined with the overall quality explanation task data to fulfill the requirements for universal knowledge acquisition. In the second stage, we train our Q-Adapt model on the entire Co-Instruct dataset.
In all stages, the first stage focuses solely on training the LoRA of the visual encoder, the language decoder, and all multimodal connector. The second stage is dedicated exclusively to training the prompt module and the multimodal connector.

\noindent\textbf{Evaluation Metric.}
For the attribute-wise perception answering task, we apply accuracy as the metric to measure the performance.
For overall quality explanation task, we adopt 5-round GPT evaluation score for comparison between our generated explanation and ground-truth explanation on completeness, precision, and relevance.
For quality assessment task, We adopt two widely used criteria for performance evaluation: Pearson linear correlation coefficient (PLCC) and Spearman rank order correlation coefficient (SROCC). A higher value for these coefficients indicates a stronger correlation with quality annotations.

\section{More Ablation Study}
%\subsection{Backbone Design Ablation}
\noindent \textbf{ The Encoder Structure for V-T Generator.} The analysis for the encoder strure of V-T Generator is shown in Table~\ref{exp_prompt2}. Utilizing the Q-Former~\citep{instructblip} can achieve an average accuracy increase of 1.43\% on Q-bench-A1 for instruction tuning on Q-Instruct, compared to the BERT~\citep{Bert} structure. It demonstrates that the Q-Former, by introducing learnable queries, can capture high-level semantic information from the instructions, facilitating the extraction of crucial task information.

%\vspace{-2mm}

\begin{table}[htbp]
\centering
\footnotesize
\caption{Comparison with different text encoders for generating instruction-guided visual prompt.}
\resizebox{0.46\textwidth}{!}{\begin{tabular}{@{}lccc|ccc@{}}
\toprule
\textbf{} & \textbf{Q-bench-A1 (dev)} & \textbf{Q-bench-A1 (test)} & \textbf{Average} & \textbf{Q-bench2-A1 (dev)} & \textbf{Q-bench2-A1 (test)} & \textbf{Average}\\ \midrule

BERT$^{Q}$          & 75.72         & 75.65       & 75.69  &   55.10         &  53.15  &  54.12         \\
\textbf{Q-Former}$^{Q}$ & \textbf{77.19} & \textbf{ 77.06}  & \textbf{77.12} &\textbf{55.80}  & \textbf{55.45}   &\textbf{55.63}  \\  \midrule
BERT$^{Co}$          &76.02          &\textbf{76.05}  &\textbf{76.12}          &  76.08         &   76.57 & 76.83         \\
\textbf{Q-Former}$^{Co}$ & \textbf{76.05} & 76.12 & 76.08 & \textbf{77.20  }   & \textbf{78.38}  &\textbf{77.79}  \\ \bottomrule
\end{tabular}}
\label{exp_prompt2}

\end{table}

\begin{figure}
\centering

\includegraphics[width=0.3\textwidth]{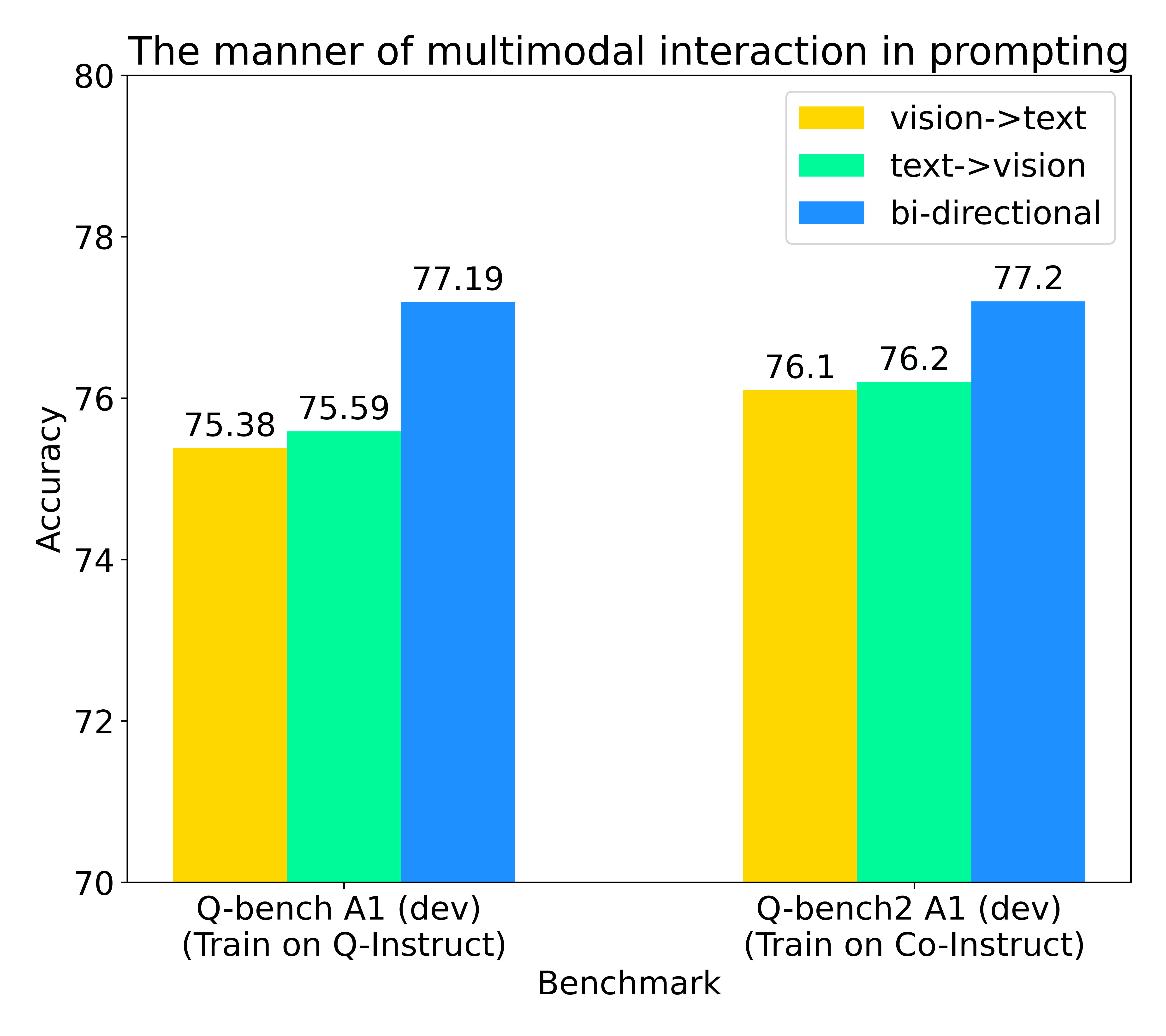} % 替换为您的图片文件

%\vspace{-10pt} % 调整图像下方的间距
\caption{The effect of variants for multimodal interaction.}
 
\label{exp_prompt3}
\end{figure}

\noindent \textbf{ Multimodal Interaction.} The multimodal interaction for constructing instruction-adaptive visual prompts is detailed in Fig.~\ref{exp_prompt3}. It can be observed that the bi-directional interaction between text and visual modalities achieves the highest performance. The performance gain from vision-text interaction (\ieno, V-T Generator) is lower than that from text-vision interaction (\ieno, T-V Prompter), which indicates the importance of mapping textual features into the visual feature space for modulating the original visual features.

\noindent \textbf{The Difference with VTC.}
VTC~\cite{VTC} concatenates the additional visual tokens to complete the original visual tokens. We conduct this insert manner like VTC to compare with our spatial-wise modulation in Table~\ref{tab:prompting_qbench1}. The results indicate that concatenating complementary visual tokens is unnecessary when using the uncompressed original visual tokens of Bunny, as the original tokens already provide sufficient information for effective processing.

\begin{table} % 'r' for right alignment, 0.5\textwidth for width
\centering
\footnotesize
\caption{Comparison of performance between our method and VTC.}

\begin{tabular}{l|c}
\toprule
\textbf{Prompting} & \textbf{Q-bench1-dev} \\ \midrule
Ours & 77.19 \\ 
VTC & 76.99 \\ \bottomrule
\end{tabular}
\vspace{-5mm}
\label{tab:prompting_qbench1}

\end{table}

\noindent \textbf{The Comparison with Co-Instruct-8B.} There are two reasons for that the performance of Q-Adapt(Co)-3B is lower than Co-Instruct-8B: \textbf{Model Scaling}: Larger models, such as the 8B parameter Co-Instruct, are inherently better at processing and leveraging larger datasets due to their greater capacity for capturing complex patterns and representations. In contrast, the smaller 3B parameter Q-Adapt may encounter limitations in handling the extensive data volume, leading to suboptimal performance. 
\textbf{Visual Token Context}: Bunny-3B has 576 visual tokens, and Co-Instruct-8B has 65 visual tokens . For multi-image tasks in supervised fine-tuning (SFT), an excessive number of visual tokens pose large challenge due to the model’s limited long contextual understanding.

In Table~\ref{tab:co_compare}, we conducted experiments on mPLUG-Owl2-8B, which serves as the backbone of Co-Instruct-8B, using a progressive tuning strategy with the efficient training method, LoRA, due to limited resources. This approach is referred to as Co-Adapt-8B-LoRA, and the results are presented in the following table. We can obtain two results: For mPLUG-Owl2, under the efficient training setting, the progressive tuning strategy achieves better performance compared to joint training (0.6950 vs. 0.6820). However, it remains significantly lower than full fine-tuning (0.7840), which can be attributed to the difficulty of fully capturing knowledge during training when using LoRA on an 8B model. This highlights the limitations of LoRA in large-scale models under efficient tuning settings, especially for downstream task adaption. Q-Adapt-3B, built upon the baseline Bunny-3B with progressive and efficient tuning strategy, demonstrates the ability to achieve performance (0.7720) comparable to Co-Instruct-8B (0.7840) under the full fine-tuning setting. This result underscores the effectiveness of our proposed method in leveraging smaller models while maintaining competitive performance.

\begin{table}[h]
    \centering
    \resizebox{0.4\textwidth}{!}{\begin{tabular}{lccc}
        \toprule
        Model & Param &  Fine-tuning Strategy & Q-Bench2-dev \\
        \midrule
        Co-Instruct-8B-LoRA & 8B & LoRA,Joint & 0.6820 \\
        Co-Adapt-8B-LoRA & 8B & LoRA,Progressive & 0.6950 \\
        Co-Instruct-8B & 8B & Full, Joint & 0.7840 \\
        Q-Adapt-3B-LoRA & 3B & LoRA, Progressive & 0.7720 \\
        \bottomrule
    \end{tabular}}
    \caption{Performance comparison of different models and training strategies on Q-Bench2-dev. }
    \label{tab:co_compare}
\end{table}

\end{document}